%% file: arxiv.tex
\documentclass[letterpaper]{article} 
\usepackage[table]{xcolor}
\usepackage{aaai2026}  
\usepackage{times}  
\usepackage{helvet}  
\usepackage{courier}  
\usepackage[hyphens]{url}  
\usepackage{graphicx} 
\usepackage{natbib}  
\usepackage{caption} 
\usepackage{algorithm}
\usepackage{algorithmic}

\usepackage{booktabs} 
\usepackage{multirow} 
\usepackage{array} 
\usepackage{makecell} 

\usepackage{newfloat}
\usepackage{listings}
\DeclareCaptionStyle{ruled}{labelfont=normalfont,labelsep=colon,strut=off} 
\floatstyle{ruled}
\newfloat{listing}{tb}{lst}{}
\floatname{listing}{Listing}
\usepackage{amsfonts}
\usepackage{amsmath}
\usepackage{amsthm}
\usepackage{booktabs}
\usepackage{subfigure}
\usepackage{bbding}
\usepackage{pifont}

\newtheorem{theorem}{Theorem}

\title{Harnessing Adaptive Topology Representations for Zero-Shot Graph Question Answering}
\author{
    Yanbin Wei\textsuperscript{\rm 1,2}
    Jiangyue Yan\textsuperscript{\rm 1},
    Kang Chun,
    Yang Chen\textsuperscript{\rm 1},
    Hua Liu\textsuperscript{\rm 1},
    James Kwok\textsuperscript{\rm 2},
    Yu Zhang\textsuperscript{\rm 1}
}
\affiliations{
    \textsuperscript{\rm 1}Southern University of Science and Technology
    \textsuperscript{\rm 2}Hong Kong University of Science and Technology

}

\usepackage{bibentry}
\begin{document}

\maketitle

\begin{abstract}
Large Multimodal Models (LMMs) have shown generalized zero-shot capabilities in diverse domain question-answering (QA) tasks, including graph QA that involves complex graph topologies. However, most current approaches use only a single type of graph representation, namely \textbf{Topology Representation Form (TRF)}, such as prompt-unified text descriptions or style-fixed visual styles. Those ``one-size-fits-all'' approaches fail to consider the specific preferences of different models or tasks, often leading to incorrect or overly long responses.
To address this, we first analyze the characteristics and weaknesses of existing TRFs, and then design a set of TRFs, denoted by $\mathcal{F}_{ZS}$, tailored to zero-shot graph QA. We then introduce a new metric, \textbf{Graph Response Efficiency (GRE)}, which measures the balance between the performance and the brevity in graph QA. Built on these, we develop the \textbf{DynamicTRF} framework, which aims to improve both the accuracy and conciseness of graph QA. To be specific, DynamicTRF first creates a \textbf{TRF Preference (TRFP)} dataset that ranks TRFs based on their GRE scores, to probe the question-specific TRF preferences. Then it trains a \textbf{TRF router} on the TRFP dataset, to adaptively assign the best TRF from $\mathcal{F}_{ZS}$ for each question during the inference.
Extensive experiments across 7 in-domain algorithmic graph QA tasks and 2 out-of-domain downstream tasks show that DynamicTRF significantly enhances the zero-shot graph QA of LMMs in terms of accuracy and brevity simultaneously. 
\end{abstract}


\section{Introduction}

Large Multimodal Models (LMMs) have served as a universal solution for zero-shot question answering (QA) across a wide range of domains, from commonsense to complex mathematical problem-solving \citep{kuang2025natural}. Nevertheless, their zero-shot versatility faces distinctive challenges when answering structured graph problems. First, comprehending the highly structured topology where each node and edge carry explicit relational semantics is difficult for LMMs, because they differ fundamentally from the free-form text or images LMM typically encounter.
Besides this structural understanding, another difficulty emerges to perform graph-based execution for specific tasks (e.g., performing Dijkstra’s algorithm \citep{dijkstra2022note} for shortest-path, Ford–Fulkerson algorithm \citep{ford1956maximal} for maximum-flow) to produce answers for the queries, which demands complex and multi-step computation. Besides, in this paper, we focus on zero-shot graph QA, where the model answers questions using only its intrinsic knowledge without task-specific fine-tuning or in-context examples. This setting amplifies the aforementioned challenges, as the LMM has neither prior exposure to the given graph nor the opportunity to learn algorithmic routines on the fly.

\begin{figure}[t]
\centering
\includegraphics[width=0.45\textwidth]{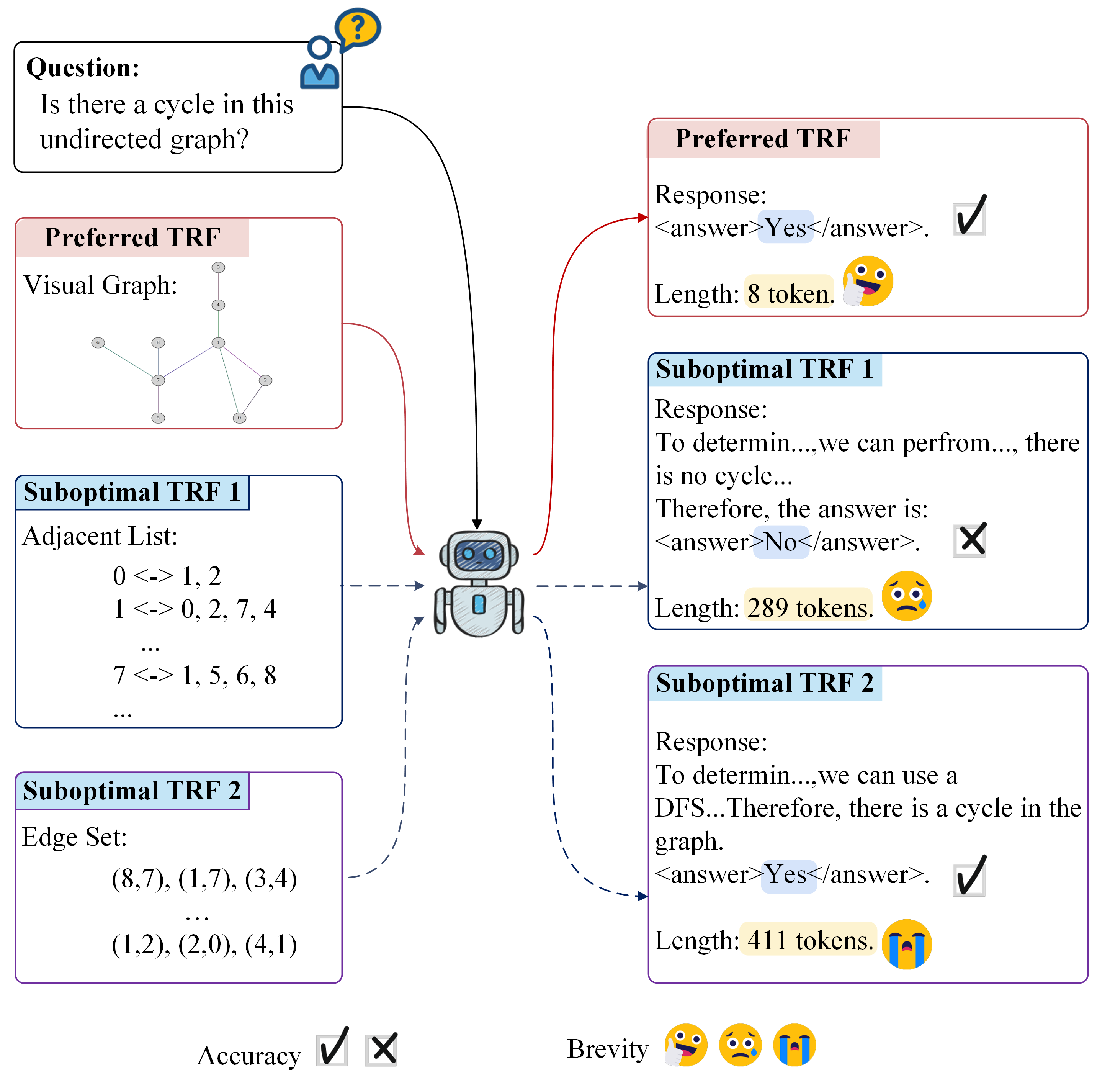}
\caption{Illustrating the impacts of diverse TRFs on the same question: suboptimal TRFs may lead to wrong answers or unduly lengthy responses compared to a preferred TRF.}
    \label{fig: intro}
\end{figure}

Existing works have adopted various topology representation forms (TRFs) to present graph topology to LMMs for understanding. Textual TRFs, as demonstrated in \citep{instructGLM, NLGraph,chen2024graphwiz, graphtoken, LLM4Graph}, encode graph structures into textual descriptions using diverse prompt templates. Recently, visual TRFs have been introduced for LMMs \citep{li2024visiongraph, wei2024gita}, which depict graph topologies as stylized images. Despite these advancements, existing methods share a common assumption: employing a single type of TRF throughout the QA process, such as unified prompt-based textual descriptions or fixed-style visualizations. This ``one-size-fits-all'' approach overlooks critical factors, including model-specific cognitive biases and task-specific representational preferences, resulting in suboptimal QA performance in specific tasks (see experiments in Table \ref{tab: ablation}). Cases in Figure \ref{fig: intro} demonstrate the QAs for the same question with diverse TRFs. As can be seen, compared to the preferred TRF, suboptimal TRFs can impede the model's comprehension of graph topology, resulting in a wrong answer or prolonged response. This raises intriguing research questions: \textit{Can these preferences for TRFs be leveraged? How and to what extent can they enhance the graph QA capabilities of LMMs?}

In this paper, we aim to address these questions through a systematic solution. Specifically, we first categorize and analyze the characteristics of existing representative TRFs. We then design and construct a TRF set \(\mathcal{F}_{ZS}\) tailored to zero-shot graph QA, under dedicated design principles.
Subsequently, we propose the \textbf{DynamicTRF} framework, which operates based on the TRFs within \(\mathcal{F}_{ZS}\) and comprises three key components: (1) defining the \textbf{Graph Response Efficiency (GRE)} metric, which balances accuracy and computational cost; (2) leveraging a probe dataset to identify TRFs within \(\mathcal{F}_{ZS}\) that have optimal GRE for each question, thereby constructing a model-specific \textbf{TRF Preference} dataset; (3) training a \textbf{TRF Router} on the TRF Preference dataset to dynamically select appropriate TRFs from \(\mathcal{F}_{ZS}\) for each question during inference. Note that since the TRFs act at the input stage of LMMs, DynamicTRF does not require access to any information about the model architecture or parameters, and thus can be applied to the state-of-the-art closed-source LMMs. 
We conduct extensive evaluations across 9 graph QA tasks, encompassing 7 in-domain algorithmic graph QA tasks and 2 out-of-domain downstream tasks. Results show that DynamicTRF enables adaptive question-specific TRF routing, balancing the QA accuracy and response brevity.

Our key contributions are summarized as follows.
\begin{itemize}
    \item We systematically investigate and discuss the existing fixed TRFs with their characteristics and limitations.
    
    \item We propose the GRE metric to quantify the accuracy-brevity trade-off for the graph QA process.
    
    \item We propose DynamicTRF, the first framework that integrates LMM graph QA with dynamic TRF routing. 
    
    \item The side-product TRFP dataset, along with our experiments, offers valuable insights into specific TRFs favored by different task categorizations.

    \item Comprehensive results show the superior performance of DynamicTRF in zero-shot graph QA across 7 in-domain and 2 out-of-domain downstream tasks.
\end{itemize}

\section{Related Work}
\label{sec: related}

\paragraph{LMM-based Graph QA.} Graph QA presents unique challenges for LMMs. These challenges stem from the need for structural awareness of graph topology and the ability to select and emulate appropriate algorithms for multi-step reasoning. Existing research in this domain can be broadly categorized into two main approaches:
(1) Toolkit-Enhanced System:
This category encompasses methods such as StructGPT \citep{structgpt}, Graph-Toolformer \citep{zhang2023graph}, and GraphDPR \citep{li2024visiongraph}. These systems utilize predefined external toolkits to assist LMMs in graph QA. Although effective within certain domains, this dependence restricts the rigid question types they can handle, rendering them inadequate for out-of-domain tasks.
(2) Graph-Aware LMMs: Examples in this category include InstructGLM \citep{tang2023graphgpt}, GraphToken \citep{graphtoken}, GraphLLM \citep{chai2023graphllm}, and Gcoder \citep{zhang2024gcoder}. These methods enhance LMMs with graph-awareness by modifying their architecture or fine-tuning their parameters, leveraging the internal knowledge of the LMMs for graph QA. Yet the additional training or architectural change breaks the zero-shot premise and makes deployment infeasible on closed-source models that provide only black-box access.



\section{Zero-shot TRF Set \(\mathcal{F}_{ZS}\)}
\label{sec: observation}
This section first categorizes and analyzes the existing representative TRFs. It then proposes important principles for designing a Zero-shot TRF Set \(\mathcal{F}_{ZS}\), and constructs an instance of \(\mathcal{F}_{ZS}\) following the proposed principles.

\subsection{Analyzing TRFs in Graph QA} 
\label{sec: analysis}
The TRFs describe the graph topology $G=\{V,E\}$, where $V$ and $E$ denote the set of nodes and edges, respectively. According to the representation types, current TRFs utilized in Graph QA can be categorized into three types: Embedding TRFs, Textual TRFs, and Visual TRFs. In Table \ref{tab: TRF_summary}, we showcase representative TRFs and summarize their characteristics.

\begin{table}[htbp]
    \centering
    \renewcommand{\arraystretch}{1.2} 
    \small 
    \scalebox{0.92}{
    \setlength{\tabcolsep}{1.5pt}
    \begin{tabular}{l|l|c|c|c|c}
        \hline
        \textbf{TRF} & \textbf{Type} & \textbf{Works} & \textbf{Charac.} & \textbf{Enc.} & \textbf{Train} \\
        \hline
        Embeddings & Embed & 
        \begin{tabular}[c]{@{}c@{}}GraphLLM\shortcite{chai2023graphllm}, \\ GraphGPT\shortcite{tang2023graphgpt}, \\ GraphToken\shortcite{graphtoken}, \\ LLaGA\shortcite{chen2024llaga}\end{tabular} & 
        \begin{tabular}[c]{@{}c@{}}Partial, \\ Informative\end{tabular} & 
        Yes & Yes \\ 
        \hline
        Visual Graph & Visual & 
        \begin{tabular}[c]{@{}c@{}}GITA\shortcite{wei2024gita}, \\ VisionGraph\shortcite{li2024visiongraph}\end{tabular} & 
        \begin{tabular}[c]{@{}c@{}}Full, \\ Intuitive, \\ Explicit\end{tabular} & 
        No & No \\
        \hline
        Edge Set & Textual & 
        \begin{tabular}[c]{@{}c@{}}GraphWiz\shortcite{chen2024graphwiz}, \\ NLGraph\shortcite{NLGraph}, \\ GPT4Graph\shortcite{guo2023gpt4graph}, \\
        GraphArena\shortcite{tang2025grapharena},
        \\
        GCoder\shortcite{zhang2024gcoder}\end{tabular} & 
        \begin{tabular}[c]{@{}c@{}}Full, \\ Sequential, \\ Implicit\end{tabular} & 
        No & No \\
        \hline
        Adjacent List & Textual & 
        \begin{tabular}[c]{@{}c@{}}InstructGLM\shortcite{instructGLM}, \\ GraphText\shortcite{zhao2023graphtext}\end{tabular} & 
        \begin{tabular}[c]{@{}c@{}}Full, \\ Sequential, \\ Implicit\end{tabular} & 
        No & No \\
        \hline
        Adjacent Matrix
        & Textual & 
        \begin{tabular}[c]{@{}c@{}}GraphDPR\shortcite{li2024visiongraph}\end{tabular} & 
        \begin{tabular}[c]{@{}c@{}}Full, \\ Redundant, \\ Implicit\end{tabular} & 
        No & No \\
        \hline
    \end{tabular}
    }
   \caption{Summary of existing TRFs. `Charac.' indicates the characteristics; `Enc.' and `Train' denote whether the TRF is generated by an external encoder and whether extra training is required, respectively. `Embed' is short for `Embedding'.}
    \label{tab: TRF_summary}
\end{table}

As the table indicates, Embedding TRFs are generated by external encoders and typically require alignment training with the embedding space of LMMs \citep{chai2023graphllm, tang2023graphgpt, chen2024llaga}. While the inevitable compression in embeddings leads to the loss of some topological information \citep{embeddingsurvey}, it also allows the learned embedding TRFs to focus on the most informative patterns.

In contrast, both Visual TRFs and Textual TRFs convey the complete topology of a graph, enabling the graph to be uniquely reconstructed from these representations. However, textual TRFs, whether in the form of an Edge Set or an Adjacency List, present complex structures in a flattened, sequential context. This makes the topological information more implicit compared to the intuitive presentation offered by visual graphs \citep{wei2024gita}. Moreover, the Adjacency Matrix often introduces redundant `0' elements for non-existent edges, especially in the case of sparse graphs, making it rarely the primary choice for graph topology representation in existing works. However, the Adjacency Matrix can still be useful in specific scenarios where the explicit representation of connections is advantageous.

\subsection{Constructing the \(\mathcal{F}_{ZS}\)}
Given the variety of available TRFs, we carefully select the most suitable ones to form a dedicated set, denoted as \(\mathcal{F}_{\text{ZS}}\), explicitly tailored to zero-shot graph QA scenarios. The selection follows three key principles:
(1) \textit{Model-Agnostic}: The generation of TRFs must be decoupled from the LMM parameters, ensuring compatibility with closed-source LMMs. This is crucial as these models often offer superior performance but do not disclose model details. \textit{Consequently, embedding-based TRFs are excluded from \(\mathcal{F}_{\text{ZS}}\) due to their reliance on embedding spaces, which are inaccessible given most leading models are closed-source}. (2)\textit{Variety}: The TRFs in \(\mathcal{F}_{ZS}\) should exhibit diversity to effectively address a wide range of question types. This diversity helps accommodate different types of graphs and QA tasks, utilizing formats like text and images that naturally align with LMM input formats. (3) \textit{Effectiveness}: Each TRF should possess strong individual capabilities, contributing significantly to the overall QA process. This ensures that each component of the set is valuable and enhances the QA performance.

Following these refined principles, we construct an instance of the zero-shot TRF set $\mathcal{F}_{ZS} = \{V_{dot}, V_{neato}, V_{circo}, V_{fdp}, V_{sfdp}, T_{set}, T_{list}, T_{mat}\}$, consisting of the following TRFs:

\begin{enumerate}
    \item \textbf{Visual TRFs (5 types):} These TRFs $V_{dot}$, $V_{neato}$, $V_{circo}$, $V_{fdp}$, and $V_{sfdp}$ are generated based on methods from \citep{wei2024gita} using different layout algorithms provided by Graphviz \citep{graphviz}. Specifically, $V_{\text{dot}}$ arranges nodes in tree-like hierarchical layers; $V_{\text{neato}}$ uses spring model \citep{fruchterman1991graph} to minimize edge crossings on canvas; $V_{\text{circo}}$ positions nodes in a circular pattern; $V_{\text{fdp}}$ offers a fast force-directed layout with optimized computational overhead; and $V_{\text{sfdp}}$ provides a scalable force-directed layout for efficiently handling large graphs.
    
    \item \textbf{Textual TRFs (3 types):} These TRFs $T_{set}$, $T_{list}$, and $T_{mat}$ present the graph topology structure via Edge set, Adjacent List, and Adjacent Matrix, respectively. The prompt templates of $T_{set}$ are retrieved from \citep{NLGraph}, and the others are designed by us.
\end{enumerate}

\begin{figure}[t]
\centering
\includegraphics[width=0.45\textwidth]{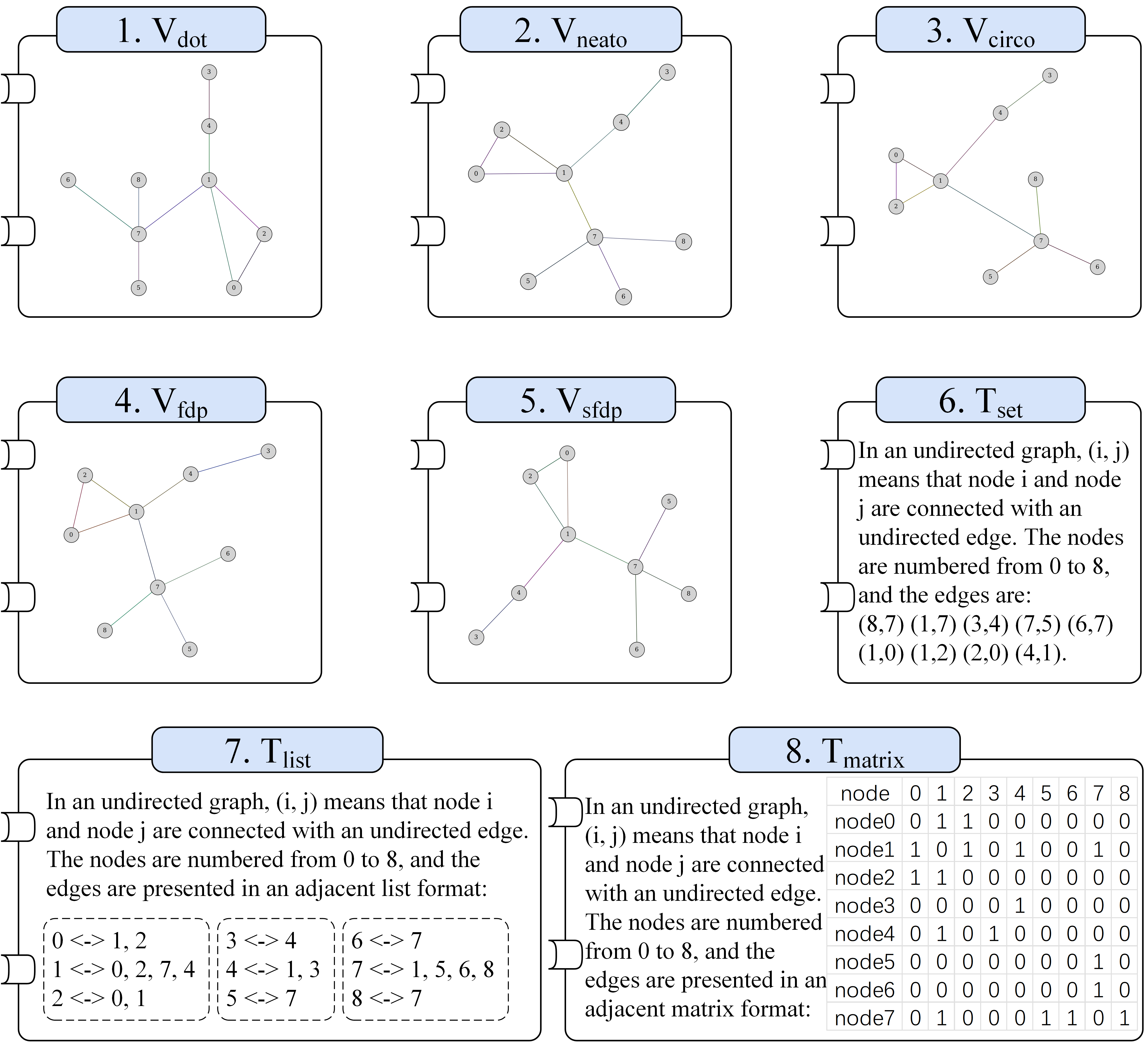}
\caption{An Illustration of eight candidate TRFs in $\mathcal{F}_{ZS}$.}
    \label{fig: examples}
\end{figure}

 We provide the TRFs generation details in Appendix A.

 Figure \ref{fig: examples} illustrates the examples of TRFs in $\mathcal{F}_{ZS}$, all of which are \textit{Model-agnostic} generated before input to LMMs, ensuring their usability with closed-source models. Visual TRFs enable rapid, intuitive perception of topology, while textual TRFs offer slower, analytical understanding, mirroring dual-system cognitive frameworks \citep{wason1974dual, daniel2017thinking}. Prior works \citep{wei2024rendering, tang2025grapharena} further show that layouts and prompt templates substantially influence graph QA performance, supporting our emphasis on \textit{Variety}. Section \ref{exp: ablation} empirically validates the \textit{effectiveness} of each TRF across tasks.

\begin{figure*}[htbp]
  \includegraphics[width=\textwidth]{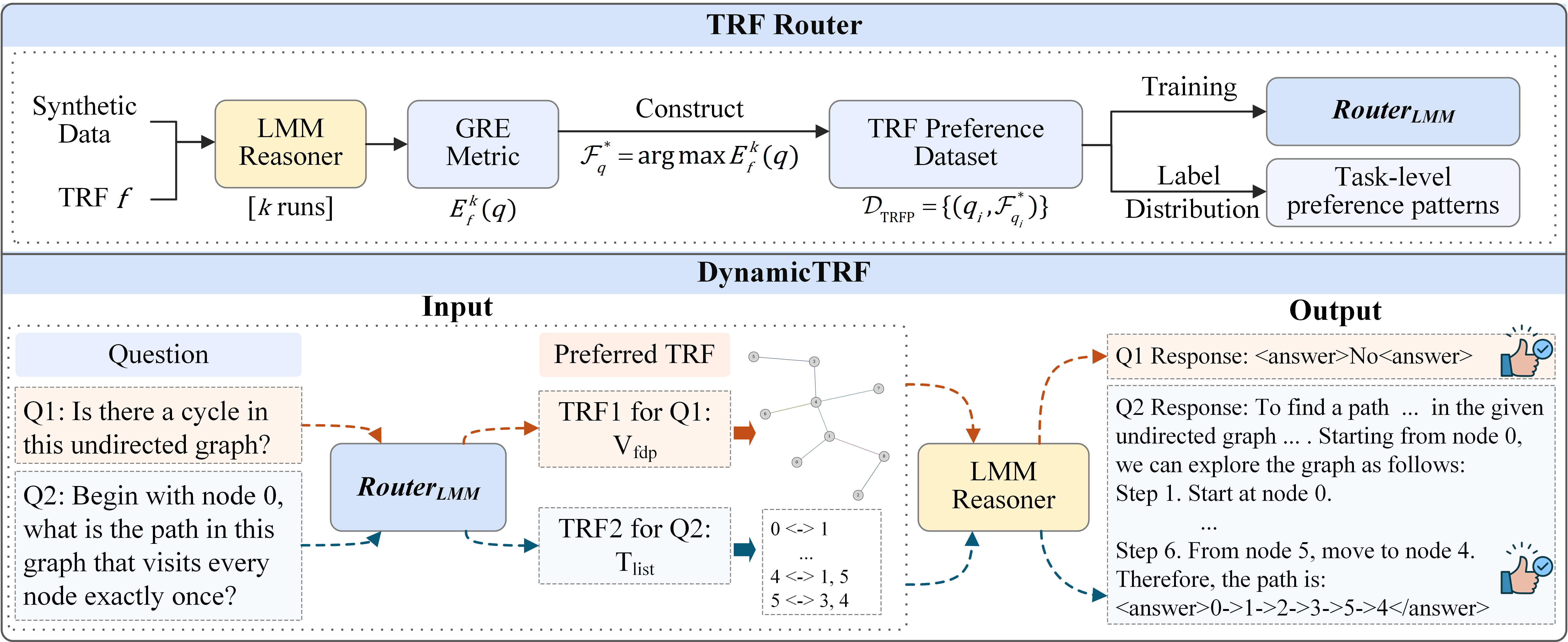}
  \caption{Overview of the DynamicTRF framework, where the TRF Router guides the LMM Reasoner to use the most appropriate TRF based on the question.}
  \label{fig:MD_framework}
\end{figure*}
\section{DynamicTRF}
\label{sec: method}
In this section, we present the DynamicTRF framework, which enhances the zero-shot graph QA of LMMs by dynamic TRF routing, on both accuracy and brevity.

\subsection{Problem Formulation}

LMM-based zero-shot graph QA leverages pretrained LMMs to tackle a wide range of graph problems, without access to or modification of the LMM specifics. This approach facilitates effective generalization across various tasks and unseen graph data, while fully preserving the inherent capabilities of LMMs across the other universal domains.

\subsection{Framework Overview}

The DynamicTRF framework, shown in Figure \ref{fig:MD_framework}, establishes a graph QA system with question-specific dynamic TRFs. It consists of two main components: (1) a LMM Reasoner for zero-shot graph QA, and (2) a TRF Router for adaptive TRF selection.

DynamicTRF defines a Graph Response Efficiency (GRE) metric to evaluate the trade-off between accuracy and computational cost of TRFs concerning the model and problem context. Using fixed probe data, DynamicTRF identifies the mapping from questions to their optimal TRFs with the highest GRE, creating a TRF Preference dataset to train the TRF Router $Router_{LMM}$.

Then, for any input question $q$, the TRF Router dynamically assigns the most suitable TRF $f_q \in \mathcal{F}_{ZS}$ and the LMM Reasoner uses $f_q$ as its TRF input to perform zero-shot inference, producing the answer $ans_q^{f_q}$. 

\subsection{Graph Response Efficiency (GRE)}
In this section, we introduce the \textit{Graph Response Efficiency (GRE)} metric, crafted to evaluate the trade-off between accuracy and computational cost among various TRFs.

For a given question $q$ and a TRF $f \in \mathcal{F}_{ZS}$ employed by the LMM Reasoner, the \textit{Graph Response Efficiency} with $k$ random runs is defined as:
\begin{equation}
\label{definition: gre}
    E_f^k(q) = \frac{100\times\text{accuracy}^k_f(q)}{(\text{avg.tok}^k_f(q))^\alpha},
\end{equation}
where $\text{accuracy}^k_f(q)$ denotes the ratio of correct answers generated by the model in $k$ runs. $\text{avg.tok}^k_f(q)$ represents the average token counts across the $k$ responses from the LMM Reasoner using TRF $f$. The hyperparameter $\alpha$ regulates the balance between accuracy and token consumption, prioritizing the importance of correct responses relative to the computational cost involved in generating them.

\subsection{TRF Preference Dataset}
\label{sec: trfp}
Using the \textit{GRE} metric, we develop the TRF Preference dataset, which explores the mapping from questions to their preferred TRFs. Specifically, we construct 7K QA pairs for in-domain tasks used in Section \ref{exp:datasets}, where the graph topology $G=(V, E)$ is randomly generated by Erdős–Rényi \citep{er_model} model ($\text{node count }N \in [3,30], \text{edge probability} \in [0.1,0.7]$) and the response correctness is validated with algorithms. For each question $q$, its preferred TRFs, denoted as $\mathcal{F}^*_q$, is determined by:
\begin{equation}
\mathcal{F}^*_q = \arg\max_{f \in \mathcal{F}_{ZS}} E^k_f(q).
\end{equation}
This equation identifies the set of TRFs $\mathcal{F}^*_q$ that maximize the GRE metric $E^k_f(q)$ for question $q$. Pairing each question $q$ with its corresponding set of preferred TRFs $\mathcal{F}^*_q$, we set $k=10$ and $\alpha = 0.5$\footnote{$\alpha$ value is user-dependent. We show its impacts in Section~\ref{exp: ablation}.} to form the TRF Preference dataset, denoted as $\mathcal{D}_{\text{TRFP}} = \{(q_i, \mathcal{F}_{q_i}^*)\}$. More details of the TRFP dataset construction method are in Appendix B.

\begin{table*}[h!tbp]
    \centering
    \setlength{\tabcolsep}{3pt}
    \renewcommand{\arraystretch}{1.3} 
    \small 
    \begin{tabular}{llccccccc}
        \hline
        Model & TRFs & Conn & Cyc & TS & SP & MF & BGM & HP \\
        \hline
        \multirow{3}{*}{\makecell[c]{GPT-4o}} & 1st 
        &\cellcolor{gray!20}$V_{fdp}$ (92.3\%) 
        &\cellcolor{gray!20} $V_{sfdp}$ (85.1\%)
        & $T_{set}$ (58.5\%)
        & $T_{set}$ (19.5\%)
        & $T_{list}$ (21.7\%)
        &\cellcolor{gray!20} $V_{dot}$ (34.2\%)
        &$T_{list}$ (30.4\%)\\
        \cline{2-9}
        & 2nd & 
        \cellcolor{gray!20}$V_{neato}$ (92.1\%) 
        & \cellcolor{gray!20}$V_{fdp}$ (12.9\%)
        & $T_{list}$ (36.2\%)
        &\cellcolor{gray!20} $V_{neato}$ (18.7\%)
        & $T_{set}$ (20.8\%)
        &\cellcolor{gray!20} $V_{circo}$ (20.3\%)
        & $T_{set}$ (20.3\%)\\ 
        \cline{2-9}
        & 3rd 
        &\cellcolor{gray!20} $V_{sfdp}$ (91.7\%)
        &\cellcolor{gray!20} $V_{dot}$ (12.2\%)
        &\cellcolor{gray!20} $V_{dot}$ (23.1\%)& $T_{list}$ (17.1\%)& $T_{mat}$ (16.7\%)
        &\cellcolor{gray!20} $V_{neato}$ (19.8\%)
        &\cellcolor{gray!20} $V_{circo}$ (18.8\%)\\
        \hline

        \multirow{3}{*}{\makecell[c]{Gemini \\ 2.5 Pro}} & 1st 
        &\cellcolor{gray!20} $V_{neato}$ (88.8\%)
        &\cellcolor{gray!20}
        $V_{neato}$ (97.2\%)
        &
        $T_{set}$ (41.0\%)
        &
        $T_{list}$ (48.6\%)
        &
        $T_{mat}$ (40.0\%)
        &\cellcolor{gray!20}
        $V_{fdp}$ (24.4\%)
        &
        $T_{list}$ (42.9\%)\\
        \cline{2-9}
        & 2nd &\cellcolor{gray!20}
        $V_{fdp}$ (88.2\%)
        & \cellcolor{gray!20} $V_{sfdp}$ (93.0\%)
        & $T_{list}$ (30.3\%)
        & $T_{mat}$ (37.6\%)
        & $T_{list}$ (36.0\%)
        & \cellcolor{gray!20} $V_{sfdp}$ (14.5\%)
        & $T_{set}$ (31.7\%)
        \\
        \cline{2-9}
        & 3rd 
        &\cellcolor{gray!20} $V_{sfdp}$ (88.2\%)
        & \cellcolor{gray!20} $V_{dot}$ (71.8\%)
        & $T_{mat}$ (15.4\%)
        & $T_{set}$ (26.6\%)
        & $T_{set}$ (14.0\%)
        & \cellcolor{gray!20} $V_{neato}$ (14.5\%)
        & $T_{mat}$ (30.2\%)
        \\
        \hline
    \end{tabular}
    \caption{Task-preferred Top-3 TRFs with frequency in TRFP dataset (GPT-4o). `Conn', `Cyc', `TS', `SP', `MF', `BGM', and `HP' denote connectivity, cycle detection, topological sort, shortest path, maximum flow, bipartite graph matching, and Hamilton path. By differing visual or textual TRFs with colors, \textbf{the special preference patterns of tasks are explicitly exposed}.}
    \label{tab: preference}
\end{table*}

The TRF Preference dataset is a crucial resource for examining the task-level preferences of TRFs. Analyzing the label frequency statistics for specific tasks can reveal the task-level TRF preferences. Table \ref{tab: preference} provides the top-3 TRF choices in various tasks of the TRFP dataset, with their frequency to be selected in $\mathcal{F}^*_q$ listed in parentheses.

We can conclude findings from Table \ref{tab: preference} by categorizing the 7 tasks into three types: \textbf{(1) Perceptual-Intensive Tasks}: Visual TRFs dominate tasks such as Connectivity, Cycle Detection, and Bipartite Graph Matching. These tasks typically require fast and intuitive topology-awareness, which visual representations are well-suited for.
\textbf{(2) Edge-Weighted Tasks}: Tasks involving edge weights, like Shortest Path and Maximum Flow, prefer textual TRFs. This preference indicates that textual representations are more analytical and suitable for computation-heavy processes. \textbf{(3) Ordered Decomposition Tasks}: Hamilton Path and Topological Sorting require ordered decomposition of the graph, and these tasks also prefer textual TRFs. This suggests that textual representations facilitate structured and sequential processing. We include the complete ranking of TRF labels with frequency statistics and more analysis in Appendix C.

\subsection{TRF Router}
\label{mind_router}
The TRF Router constitutes the core decision-making module that dynamically selects a proper TRF $f_q \in \mathcal{F}_{ZS}$ for each question $q$. Building on our GRE metric $E^k_f(q)$, we formalize the routing strategy through a dual-objective optimization framework that achieves Pareto optimality in accuracy-efficiency tradeoffs.

\paragraph{Pareto Optimal Routing.}
For each question $q$, we define
1) Accuracy objective: $\text{Acc}^k_f(q) = \log( 100\times\text{accuracy}^k_f(q))$. 2) Efficiency objective: $\text{Eff}^k_f(q) = -\log(\text{avg.tok}^k_f(q))$. Therefore, according to its definition (equation \ref{definition: gre}), the logrithm form of the GRE metric $\text{log}(E^k_f(q)) = \text{Acc}^k_f(q) + \alpha\text{Eff}^k_f(q)$. Both objectives become better when they are larger. Taking the logarithm does not change the relative size of numerical values, we have a routing strategy $R  \succ R'$ if:
\begin{equation}
    \begin{cases}
    \mathbb{E}_q[\text{Acc}^k_{f_q^R}(q)] \geq \mathbb{E}_q[\text{Acc}^k_{f_q^{R'}}(q)] \\
    \mathbb{E}_q[\text{Eff}^k_{f_q^R}(q)] \geq \mathbb{E}_q[\text{Eff}^k_{f^{R'}_q}(q)],
    \end{cases}
\end{equation}
with strict inequality in at least one objective, where $f^R_q$ and $f^{R'}_q$ are TRFs selected by $R$ and $R'$ for $q$, respectively. 

Based on this definition, we demonstrate the Pareto optimality of the optimal dynamic routing beyond all the individual TRFs by providing a detailed theorem with proof in Appendix D.
\begin{theorem}[GRE-based Dynamic Routing Pareto Optimality]
For any question distribution $\mathcal{D}_q$ and tradeoff parameter $\alpha>0$, the optimal router $R^*$ satisfying
$\forall f\in\mathcal{F}_{ZS}, R^*\succ R_f$, where $R^*$ always select a TRF $f_q^{R^*} \in \mathcal{F}^*_q$, and  $R_f$ represents the routing always select $f$.
The strict inequality establishes when $f$ is suboptimal for any $q\in\text{supp}(\mathcal{D}_q)$.
\end{theorem}

\paragraph{Router Training.}
We treat the TRF Router as a classification model (default DeBERTaV3-base) $\text{R}_\phi(q): \mathcal{Q}\mapsto\mathcal{F}_{ZS}$ and train it using the TRFP dataset $\mathcal{D}_{\text{TRFP}} = \{(q_i, F^*_{q_i})\}$. For each TRF $f \in \mathcal{F}_{ZS}$, we define $y_f$ as an indicator of whether $f$ is in the true label set $F^*_q$:
\( 
y_f = \mathbb{I}[f \in F^*_q] 
\)
The loss function is then defined as:
\begin{align*}
\mathcal{L}(\phi) = -\mathbb{E}_{(q,F^*_q)\sim\mathcal{D}_{\text{TRFP}}} \Bigg[ & \sum_{f \in \mathcal{F}_{ZS}} y_f \log p_\phi(y_f|q) \\
& + (1 - y_f) \log (1 - p_\phi(y_f|q)) \Bigg],
\end{align*}
where $p_\phi(y_f|q)$ represents the probability that TRF $f$ is present in the true label set $F^*_q$. This formulation allows the router to approximate the optimal $R^*$. 

\section{Experiment}
In this section, we empirically evaluate the proposed DynamicTRF framework.

\begin{table*}[h!tbp]
    \centering
    \setlength{\tabcolsep}{0.3pt}
    \renewcommand{\arraystretch}{1.25} 
    \small 
    \begin{tabular}{cccccccccccccccc}
        \hline
        & \multicolumn{2}{c}{\raisebox{-0.6ex}[0pt]{Conn}} 
        & \multicolumn{2}{c}{\raisebox{-0.6ex}[0pt]{Cyc}} 
        & \multicolumn{2}{c}{\raisebox{-0.6ex}[0pt]{TS}} 
        & \multicolumn{2}{c}{\raisebox{-0.6ex}[0pt]{SP}} 
        & \multicolumn{2}{c}{\raisebox{-0.6ex}[0pt]{MF}} 
        & \multicolumn{2}{c}{\raisebox{-0.6ex}[0pt]{BGM}} 
        & \multicolumn{2}{c}{\raisebox{-0.6ex}[0pt]{HP}} \\
        Method
        &\multicolumn{2}{c}{\raisebox{0.5ex}{\rule{2.05cm}{0.4pt}}}
        &\multicolumn{2}{c}{\raisebox{0.5ex}{\rule{2.05cm}{0.4pt}}}
        &\multicolumn{2}{c}{\raisebox{0.5ex}{\rule{2.05cm}{0.4pt}}}
        &\multicolumn{2}{c}{\raisebox{0.5ex}{\rule{2.05cm}{0.4pt}}}
        &\multicolumn{2}{c}{\raisebox{0.5ex}{\rule{2.05cm}{0.4pt}}}
        &\multicolumn{2}{c}{\raisebox{0.5ex}{\rule{2.05cm}{0.4pt}}}
        &\multicolumn{2}{c}{\raisebox{0.5ex}{\rule{2.05cm}{0.4pt}}}
        \\
         & Acc(Tok) & GRE & Acc(Tok) & GRE & Acc(Tok) & GRE & Acc(Tok) & GRE & Acc(Tok) & GRE & Acc(Tok) & GRE & Acc(Tok) & GRE \\
        \midrule
        \multicolumn{15}{c}{\textit{GPT-4o}} \\
        \midrule
        CoT 
        & 92.5(\underline{273.3}) & \underline{5.6}
        & 52.7(480.6) & 2.4
        & 36.6(224.2) & 2.4
        & 54.6(566.0) & 2.3
        & 25.3(362.9) & 1.3
        & 69.5(370.1) & 3.6
        & 50.0(\underline{124.9}) & 4.5
        \\
        NLGraph 
        & 92.9(296.6)& 5.4
        & 60.2(337.9)& 3.3 
        & 36.2(\underline{202.6})& 2.5
        & 59.0(533.7)& 2.6
        & 25.6(\textbf{335.1})& 1.4 
        & 62.2(\underline{356.3})& 3.3
        & 57.1(176.5)& 4.3
        \\
        GraphDPR
        & \underline{94.3}(412.2)& 4.7
        & \underline{68.7}(626.4)& 2.7
        & 40.2(411.0)& 2.0
        & \underline{59.1}(\underline{496.3})& \underline{2.7}
        & \underline{31.2}(502.7)& \underline{1.4}
        & 76.5(582.1)& 3.2
        & \textbf{62.7}(475.8)& 2.9
        \\
        GITA
        & 93.4(285.3)& 5.5
        & 64.7(\underline{325.7})& \underline{3.6}
        & \textbf{40.9}(256.1)& \underline{2.6}
        & 56.8(\textbf{482.1})& 2.6
        & 27.4(\underline{359.9})& 1.4
        & \underline{82.5}(392.4)& \underline{4.2}
        & 54.9(188.8)& 4.0
        \\
        \rowcolor{gray!20} 
        DynamicTRF
        & \textbf{96.1}(\textbf{38.8}) & \textbf{15.4} 
        & \textbf{89.3}(\textbf{75.9})& \textbf{10.3}
        & \underline{40.8}(\textbf{176.1})& \textbf{3.1}
        & \textbf{68.4}(499.1)& \textbf{3.1}
        & \textbf{36.6}(385.2)& \textbf{1.9}
        & \textbf{92.0}(\textbf{233.6})& \textbf{6.0}
        & \underline{61.1}(\textbf{76.3})& \textbf{7.0}
        \\
        \midrule
        \multicolumn{15}{c}{\textit{Gemini-2.5 Pro}} \\
        \midrule
        CoT
        &97.2(\underline{218.7}) & 6.2
        &98.6(716.6) & 3.7
        &84.1(\underline{1395.9}) & 2.3
        &93.6(810.8) & 3.3
        &91.7(1154.9) & 2.7
        &97.1(1076.8) & 3.0
        &96.8(\underline{678.1}) & 3.7
        \\
        NLGraph 
        & 97.3(220.0) & \underline{6.5}
        & 98.6(629.9) & 3.9
        & 84.5(1437.8) & 2.2
        & 94.0(847.5) & 3.2
        & 92.2(\underline{1062.9}) & \underline{2.8}
        & 97.1(\underline{1026.4}) & \underline{3.0}
        & 97.1(742.9) & 3.6
        \\
        GraphDPR
        & \underline{98.5}(396.6) & 4.9
        & 98.9(907.7) & 3.3
        & \underline{86.2}(1651.4) & 2.1
        & \underline{95.5}(849.5) & 3.3
        & \underline{94.8}(1319.6) & 2.6
        & 97.8(1187.2) & 2.8
        & \underline{97.5}(992.9) & 3.1
        \\
        GITA
        & 98.2(238.5) & 4.9
        & \underline{99.2}(\underline{478.9}) & \underline{4.5} 
        & 85.5(1398.0) & \underline{2.3}
        & 94.3(\underline{808.2}) & \underline{3.3}
        & 94.2(1155.1) & 2.8
        & \underline{99.1}(1078.2) & 3.0
        & 97.3(679.8) & \underline{3.7}
        \\
        \rowcolor{gray!20} 
        DynamicTRF
        &\textbf{100}(\textbf{12.9}) & \textbf{27.8}
        &\textbf{99.3}(\textbf{16.7}) & \textbf{24.3}
        &\textbf{87.8}(\textbf{1191.2}) & \textbf{2.5}
        &\textbf{96.3}(\textbf{798.6}) & \textbf{3.4}
        &\textbf{100}(\textbf{1004.8}) & \textbf{3.2}
        &\textbf{100}(\textbf{776.0}) & \textbf{3.6}
        &\textbf{100}(\textbf{254.6}) & \textbf{6.3}
        \\
        
        \hline
    \end{tabular}
    \caption{Zero-shot capabilities on in-domain graph algorithmic tasks.  Acc, Tok, and GRE refer to task-average accuracy (\%), token consumption, and GRE, respectively.}
    \label{tab: main1}
\end{table*}

\subsection{Experimental Setup}
\label{exp:setup}
\paragraph{Datasets.}
\label{exp:datasets}
We evaluate DynamicTRF on both seven in-domain graph algorithmic tasks and two out-of-domain downstream application tasks under a zero-shot setting. The in-domain datas are retrieved from GVLQA-BASE benchmark \citep{wei2024gita}, containing tasks including identifying Connectivity \citep{sedgewick2001algorithms}, Cycle \citep{sedgewick2001algorithms}, and computing Topological Sorting \citep{kahn1962topological}, Shortest Path \citep{dijkstra2022note}, Maximum Flow \citep{ford1956maximal}, Bipartite Graph Matching \citep{karp1990optimal}, and Hamilton Path \citep{gould2003advances} (denoted as 'Conn', 'Cyc', 'TS', 'SP', 'MF', 'BGM', and 'HP', respectively). For out-of-domain applications, we adopt the ca-GrQC and ca-HepTh \citep{cadataset} datasets for link prediction (LP), and use the PolBlog \citep{adamic2005political} and Cora \citep{yang2016revisiting} datasets for the node classification (NC) task. Data and task details are in Appendix E. 

\paragraph{Baselines.}We evaluate the zero-shot graph QA capabilities of DynamicTRF against the following baselines for in-domain tasks: (1) Vanilla Chain-of-Thought (CoT) \citep{wei2022chain}, which uses step-by-step prompts; 
(2) NLGraph \citep{NLGraph} that utilizes BAG prompting to conceptualize a graph and algorithmic prompting to specify the algorithm;
(3) GraphDPR \citep{li2024visiongraph} employs external tools to generate intermediate descriptions and code with multi-step reasoning;
(4) GITA \citep{wei2024gita} that pairs a visual TRF and a textual TRF to collaboratively solve graph algorithmic problems. For monetary cost, we only evaluate all methods with LMMs GPT-4o \citep{gpt4o} and Gemini-2.5 Pro \citep{gemini2.5}. For each question, we evaluate $k=3$ times with temperature $\tau=0.7$. We report the task-average accuracy (noted Acc\%) and token cost (noted Tok) across trials, and calculate task-average $GRE = Acc/Tok^\alpha$, $\alpha =0.5$\footnote{GRE scores averaged on questions are not adopted as metrics since susceptible to outliers.}. For out-of-domain tasks, we use the base model, CoT, and GITA as baselines\footnote{Methods like NLGraph and GraphDPR are not applicable due to the absence of definitive algorithms for downstream tasks.}. Implementation details of all methods are in Appendix F.

\subsection{Main Results}
Tables \ref{tab: main1} and \ref{tab: main2} compare the effectiveness (accuracy), efficiency (token consumption), and the trade-off metric GRE of DynamicTRF against baseline methods across seven in-domain and two out-of-domain tasks. DynamicTRF consistently outperforms the baselines by effectively balancing accuracy and efficiency, achieving the highest GRE metric across all models and tasks.

A closer examination of accuracy and token consumption across task types reveals further insights: (1) For perceptual-intensive tasks (Conn, Cyc, BGM), DynamicTRF significantly enhances accuracy while reducing token consumption. These tasks demand rapid and intuitive perception, favoring visual TRFs that excel in both accuracy and token efficiency, thus enabling effective and swift reasoning. (2) In edge-weighted tasks (SP, MF), accuracy is prioritized over token consumption in the trade-off. This is because the routing tends to select TRFs suited for analytical computation, which also incur higher token costs. (3) For ordered decomposition tasks (TS, HP), the trade-off favors token consumption over accuracy. Specifically, in these tasks, the less prevalent TRFs selected in rare cases are observed to contribute a lot to token savings, revealing the diverse sample-wise demands within these tasks.  (4) Despite the out-of-domain nature of downstream applications (LP, NC), the router consistently improves both accuracy and token consumption. This suggests that the advantages of the TRF router can be effectively transferred across domains, enhancing DynamicTRF's potential as a robust and general solution for improving zero-shot capabilities.
\begin{table}[tbp]
    \centering
    \setlength{\tabcolsep}{3.5pt}
    \renewcommand{\arraystretch}{1.25} 
    \small 
    \begin{tabular}{lcccc}
        \hline
        & \multicolumn{2}{c}{\raisebox{-0.6ex}[0pt]{LP}} 
        & \multicolumn{2}{c}{\raisebox{-0.6ex}[0pt]{NC}} 
       \\
        Method
        &\multicolumn{2}{c}{\raisebox{0.5ex}{\rule{2.05cm}{0.4pt}}}
        &\multicolumn{2}{c}{\raisebox{0.5ex}{\rule{2.05cm}{0.4pt}}}
        
        \\
         & Acc(Tok) & GRE & Acc(Tok) & GRE \\
        \midrule
        GPT-4o
        & 71.8(\underline{210.5}) & 5.0
        &55.7(242.1) & 3.6
        \\ \hline
         \ +CoT
         & 73.2(256.3) & 4.6
         &56.8(291.7) & 3.3
        \\
         \ +GITA
         & \underline{78.1}(224.5) & \underline{5.2}
         &\underline{60.2}(\underline{205.5}) & \underline{4.2}
        \\ 
         \rowcolor{gray!20}\ +DynamicTRF
         & \textbf{81.4}(\textbf{163.3}) & \textbf{6.5}
        & \textbf{68.2}(\textbf{126.1}) & \textbf{5.9} \\ 
        \midrule
        Gemini-2.5 Pro
        &77.2(\underline{300.1}) & 4.3
        &60.5(330.5) & 3.4
        \\ \hline
         \ +CoT
         &77.5(330.5) & 4.2
         &60.9(360.8) & 3.1
        \\
         \ +GITA 
         &\underline{80.3}(310.2) & \underline{4.5}
         &\underline{62.8}(\underline{302.4}) & \underline{3.7}
        \\ 
         \rowcolor{gray!20}\ +DynamicTRF
         &\textbf{83.8}(\textbf{225.7}) & \textbf{5.6}
        &\textbf{72.5}(\textbf{183.6}) & \textbf{5.4} \\
        \hline
    \end{tabular}
    \caption{Zero-shot capacities on out-of-domain tasks.}
    \label{tab: main2}
\end{table}
\begin{table*}[t]
    \centering
    \setlength{\tabcolsep}{0.3pt}
    \renewcommand{\arraystretch}{1.1} 
    \small 
    \begin{tabular}{lccccccccccccccc}
        \hline
        & \multicolumn{2}{c}{\raisebox{-0.6ex}[0pt]{Conn}} 
        & \multicolumn{2}{c}{\raisebox{-0.6ex}[0pt]{Cyc}} 
        & \multicolumn{2}{c}{\raisebox{-0.6ex}[0pt]{TS}} 
        & \multicolumn{2}{c}{\raisebox{-0.6ex}[0pt]{SP}} 
        & \multicolumn{2}{c}{\raisebox{-0.6ex}[0pt]{MF}} 
        & \multicolumn{2}{c}{\raisebox{-0.6ex}[0pt]{BGM}} 
        & \multicolumn{2}{c}{\raisebox{-0.6ex}[0pt]{HP}} \\
        TRF
        &\multicolumn{2}{c}{\raisebox{0.5ex}{\rule{2.05cm}{0.4pt}}}
        &\multicolumn{2}{c}{\raisebox{0.5ex}{\rule{2.05cm}{0.4pt}}}
        &\multicolumn{2}{c}{\raisebox{0.5ex}{\rule{2.05cm}{0.4pt}}}
        &\multicolumn{2}{c}{\raisebox{0.5ex}{\rule{2.05cm}{0.4pt}}}
        &\multicolumn{2}{c}{\raisebox{0.5ex}{\rule{2.05cm}{0.4pt}}}
        &\multicolumn{2}{c}{\raisebox{0.5ex}{\rule{2.05cm}{0.4pt}}}
        &\multicolumn{2}{c}{\raisebox{0.5ex}{\rule{2.05cm}{0.4pt}}}
        \\
                 & Acc(Tok) & GRE & Acc(Tok) & GRE & Acc(Tok) & GRE & Acc(Tok) & GRE & Acc(Tok) & GRE & Acc(Tok) & GRE & Acc(Tok) & GRE \\
        \midrule
        \multicolumn{15}{c}{\textit{GPT-4o}} \\
        \midrule
        $V_{dot}$
        &78.5(8.4) & 27.1
        &80.0(\underline{8.0}) & {28.3}
        &12.3(346.0) & 0.7
        &14.5(\underline{146.3}) & 1.2
        &7.5(410.2) & 0.4
        &\underline{91.2}(\underline{244.3}) & \underline{5.8}
        &13.3(\textbf{40.0}) & 2.1
        \\
        $V_{neato}$
        &95.1(\textbf{7.7}) & \textbf{34.3}
        &\underline{87.1}(\underline{8.0}) & \underline{30.8}
        &3.0(\textbf{30.0}) & 0.5
        &20.0(\textbf{130.1}) & 1.8
        &10.8(387.4) & 0.5
        &87.2(280.0) & 5.2
        &11.1(\textbf{40.0}) & 1.8
        \\
        $V_{circo}$
        &89.7(8.2) & 31.3
        &70.7(\underline{8.0}) & 25.0
        &3.0(\textbf{30.0}) & 0.5
        &18.2(153.5) & 1.5
        &8.1(421.3) & 0.4
        &88.2(271.6) & 5.3
        &14.4(\textbf{40.0}) & 2.3
        \\
        $V_{fdp}$
        &\underline{96.0}(\underline{8.1}) & \underline{33.6}
        &60.5(8.0) & 21.4
        &3.0(\textbf{30.0}) & 0.5
        &17.3(150.5) & 1.4
        &8.1(389.2) & 0.4
        &82.9(295.3) & 4.8
        &4.4(74.0) & 0.5
        \\
        $V_{sfdp}$
        &94.8(\underline{8.1}) & 33.3
        &85.1(\textbf{7.0}) & \textbf{32.2}
        &7.0(120.0) & 0.6
        &25.5(168.7) & 2.0
        &8.1(386.1) & 0.4
        &84.0(302.9) & 4.8
        &12.2(40.0) & 1.9
        \\
        $T_{set}$
        & 92.5(273.3) & 5.6
        & 52.7(480.6) & 2.4
        & \underline{36.6}(224.2) & \underline{2.4}
        & 54.6(566.0) & 2.3
        & \underline{25.3}(\textbf{362.9}) & \underline{1.3}
        & 69.5(370.1) & 3.6
        & \underline{50.0}(124.9) & 4.5
        \\
        $T_{list}$ 
        & 89.0(218.2) & 6.0
        & 53.2(359.7) & 2.8
        & 34.2(206.3) & 2.4
        & \underline{55.5}(518.9) & \underline{2.4}
        & 24.2(400.7) & 1.2
        & 65.8(410.8) & 3.2
        & 50.0(107.0) & \underline{4.8}
        \\
        $T_{mat}$
        &79.9(233.7) & 5.2
        &52.7(417.1) & 2.6
        &6.8(378.1) & 0.4
        &34.5(591.8) & 1.4
        &17.2(402.9) & 0.9
        &60.4(407.3) & 3.0
        &31.1(160.9) & 2.5
        \\
        \rowcolor{gray!20}  TRF Router
        & \textbf{96.1}(38.8) & 15.4 
        & \textbf{89.3}(75.9)& 10.3
        & \textbf{41.4}(176.1)& \textbf{3.1}
        & \textbf{68.4}(499.1)& \textbf{3.1}
        & \textbf{36.6}(\underline{385.2})& \textbf{1.9}
        & \textbf{92.0}(\textbf{233.6})& \textbf{6.0}
        & \textbf{61.1}(76.3)& \textbf{7.0}
        \\
        \midrule
        \textit{Ideal Routing}
        &\textit{100(7.9)} & \textit{35.6}
        &\textit{100(7.1)} & \textit{37.6}
        &\textit{44.5(268.0)} & \textit{2.7}
        &\textit{81.8(223.4)} & \textit{5.5}
        &\textit{53.8(380.0)} & \textit{2.8}
        &\textit{100(181.7)} & \textit{7.4}
        &\textit{76.7(72.2)} & \textit{9.0}
        \\
        \hline
    \end{tabular}
    
    \caption{Comparison of the DynamicTRF framework with TRF Router versus single TRFs on in-domain tasks. `\textit{Ideal Routing}' means the routing is always to the best TRF. We leave Gemini-2.5 Pro results in Appendix G due to page limit.}
    \label{tab: ablation}
\end{table*}

\subsection{Ablation Study}
\label{exp: ablation}
\paragraph{Routing Necessity.} To elucidate the necessity of the proposed TRF Routing, we present metrics for individual TRFs in Table \ref{tab: ablation}, compared with the proposed TRF Router within the DynamicTRF framework. We also provide performance of \textit{`Ideal Routing'}, where the TRF with the optimal GRE is always selected, serving as the upper bound capability of TRF routing. As can be seen, there does not exist any TRF that can dominate the others across tasks. For each task, the top TRFs are highly aligned with the task preferences in the TRFP dataset (Table \ref{tab: preference}). With the TRF Router, DynamicTRF achieves the highest GRE scores across all tasks beyond all individual TRFs, underscoring its superior performance in balancing accuracy and efficiency. Besides, there is still a gap between the TRF Router and `Ideal Routing', highlighting the preserved promising potential of TRF routing. 

\paragraph{Sensitivity on  $\alpha$.}
Table \ref{tab: alpha} demonstrates the performance effects of varying $\alpha$ values in Eq.(\ref{definition: gre}), where the routers are separately trained on diverse TRFP datasets with updated GRE metrics computed with $\alpha=0$ and $1$. As illustrated, setting a lower $\alpha=0$ makes GRE dominant by accuracy, thus increasing the accuracy but costing more tokens. Conversely, a higher $\alpha=1$ value penalizes more on token consumption, resulting in reduced tokens but decreased accuracy.

\begin{table}[h!tbp]
    \centering
    \small
    \setlength{\tabcolsep}{8pt}
    \renewcommand{\arraystretch}{0.78} 
        \begin{tabular}{lccccc}
            \toprule
            \textbf{Metric} 
            & $\alpha$ 
            & I.D. 
            & LP 
            & NC 
            & \textbf{Avg.} \\
            \midrule
            \multicolumn{6}{c}{\textit{GPT-4o}} \\
            \midrule
            
            \multirow{2}{*}{$\Delta \text{Acc} \uparrow$}
            & 0 & +0.0 & +0.7 & +0.5 & +0.4  \\
            & 1 & -0.1 & -0.4 & -0.2 & -0.2 \\

            \midrule
            
            \multirow{2}{*}{$\Delta \text{Tok} \downarrow$}
            &  0 & +78.8 & +45.6 & +32.7 & +52.4  \\
            &  1 & -2.1 & -3.5 & -2.2 & -2.6 \\
            \midrule
            \multicolumn{6}{c}{\textit{Gemini-2.5 Pro}} \\
            \midrule
            
            \multirow{2}{*}{$\Delta \text{Acc} \uparrow$}
            &  0 & +0.7 & +1.2 & +2.5 & +1.5  \\
            &  1 & -19.2 & -12.1 & -6.9 & -12.7 \\
            \midrule
            
            \multirow{2}{*}{$\Delta \text{Tok} \downarrow$}
            & 0 & +286.2 & +72.9 & +83.8 & +147.6  \\
            & 1 & -8.5 & -4.2 & -5.7 & -6.1  \\
            \bottomrule
        \end{tabular}
        \caption{The performance changes with varying the $\alpha$ value from $0.5$ to $0$/$1$. `I.D.' means an average on in-domain tasks.}
        \label{tab: alpha}
\end{table}

\subsection{Router Transferability for LMMs}
\begin{table}[htbp]
    \centering
    \small
    \setlength{\tabcolsep}{1.8pt}
    \renewcommand{\arraystretch}{0.8} 
        \begin{tabular}{lcccc}
            \toprule
            \textbf{Metric} 
            & I.D. 
            & LP 
            & NC 
            & \textbf{Avg.} \\
            \midrule
            \multicolumn{5}{c}{\textit{Transfer Gemini-2.5 Pro's Router to GPT-4o}} \\
            \midrule
            
            $\Delta \text{Acc} \uparrow$
             & -0.3(+6.1) & -0.1(+3.2) & -1.4(+6.6) & -0.6(+5.3)  \\

            \midrule
            
            $\Delta \text{Tok} \downarrow$
             & -35.5(-123.4) & -9.2(-56.4) & -40.1(-119.5) & -28.3(-99.8)   \\
            \midrule
            \multicolumn{5}{c}{\textit{Transfer GPT-4o's Router to Gemini-2.5 Pro}} \\
            \midrule
            
            $\Delta \text{Acc} \uparrow$
            & +0.1(+1.9) & +0.5(+4.0) & -0.1(+9.6) & +0.2(+5.2)  \\
            \midrule
            
            $\Delta \text{Tok} \downarrow$     & +14.4(-216.2) & +26.9(-47.5) & +3.2(-115.6) & +14.8(-126.4) \\
            \bottomrule
        \end{tabular}

\caption{Performance changes when the LMM Reasoner uses a transferred router from other LMMs in DynamicTRF. Values in parentheses are relative to the best baselines.}
\label{tab: transfer}
\end{table}

To evaluate the cross-model generalization of the proposed method, we conducted transfer experiments where the router trained on one LMM (either GPT-4o or Gemini-2.5 Pro) is directly used by the other without retraining. This setup tests whether TRF preferences learned from one model's behavior could be effectively leveraged by another. As shown in Table \ref{tab: transfer}, when GPT-4o uses the Gemini-2.5 Pro's Router, the token count of responses decreases, but the accuracy drops, while using GPT-4o's router in Gemini-2.5 Pro has the opposite effect. \textbf{Anyway, the transferred routers consistently provided benefits} beyond the baselines in terms of both accuracy and brevity. This indicates that the learned TRF preferences show promise in transferring beyond model biases.

\section{Conclusion}
We propose DynamicTRF, a novel framework to enhance zero-shot graph QA in LMMs. By systematically investigating existing TRF characteristics and proposing the GRE metric, we explicitly observe the diverse preferences on TRFs in graph QAs and highlight the necessity to utilize them. By dynamically assigning proper TRF for queries, DynamicTRF achieves a remarkable balance between accuracy and response brevity without model modification. Extensive experiments across 7 in-domain algorithmic and 2 out-of-domain downstream tasks demonstrate the practicality of DynamicTRF in zero-shot graph QA.

\bibliography{main}

\clearpage
\input{Appendix}

\end{document}

%% file: Appendix.tex
\appendix

\renewcommand{\thetable}{A\arabic{table}}
\renewcommand{\thefigure}{A\arabic{figure}}


\onecolumn

\section{A. TRF Generation}
In this section, we provide more detailed introductions for the TRFs in $\mathcal{F}_{ZS}$ and their generation approaches. As introduced, $\mathcal{F}_{ZS}$ contains 8 types of dedicated TRFs, where $V_{dot}, V_{neato}, V_{circo}, V_{fdp}$ and $V_{sfdp}$ are visual TRFs, and $ T_{set}, T_{list}$, and $T_{mat}$ are textual TRFs.  

\paragraph{More Details for Visual TRFs Generation}
These visual TRFs are generated by leveraging diverse layout algorithms to compute graph layouts displayed on the canvas, specifically $V_{dot}$, $V_{neato}$, $V_{circo}$, $V_{fdp}$, and $V_{sfdp}$. Their generation follows the methodologies outlined in \citep{wei2024gita} and \citep{wei2024rendering}, using the graph visualization tool Graphviz \citep{graphviz}. To ensure style consistency without unnecessary disturbances, we only vary the layout algorithms while fixing other configurations (white background, circular node shape (except for barpartite graph matching, we use rectangular/circular shape to distinguish the hosts/tasks), default edge thickness) in Graphviz during generation. Specifically, the five visual TRFs, each utilizing a distinct layout algorithm, are introduced as follows:
\begin{figure}[htbp]
\centering
\includegraphics[width=0.6\textwidth]{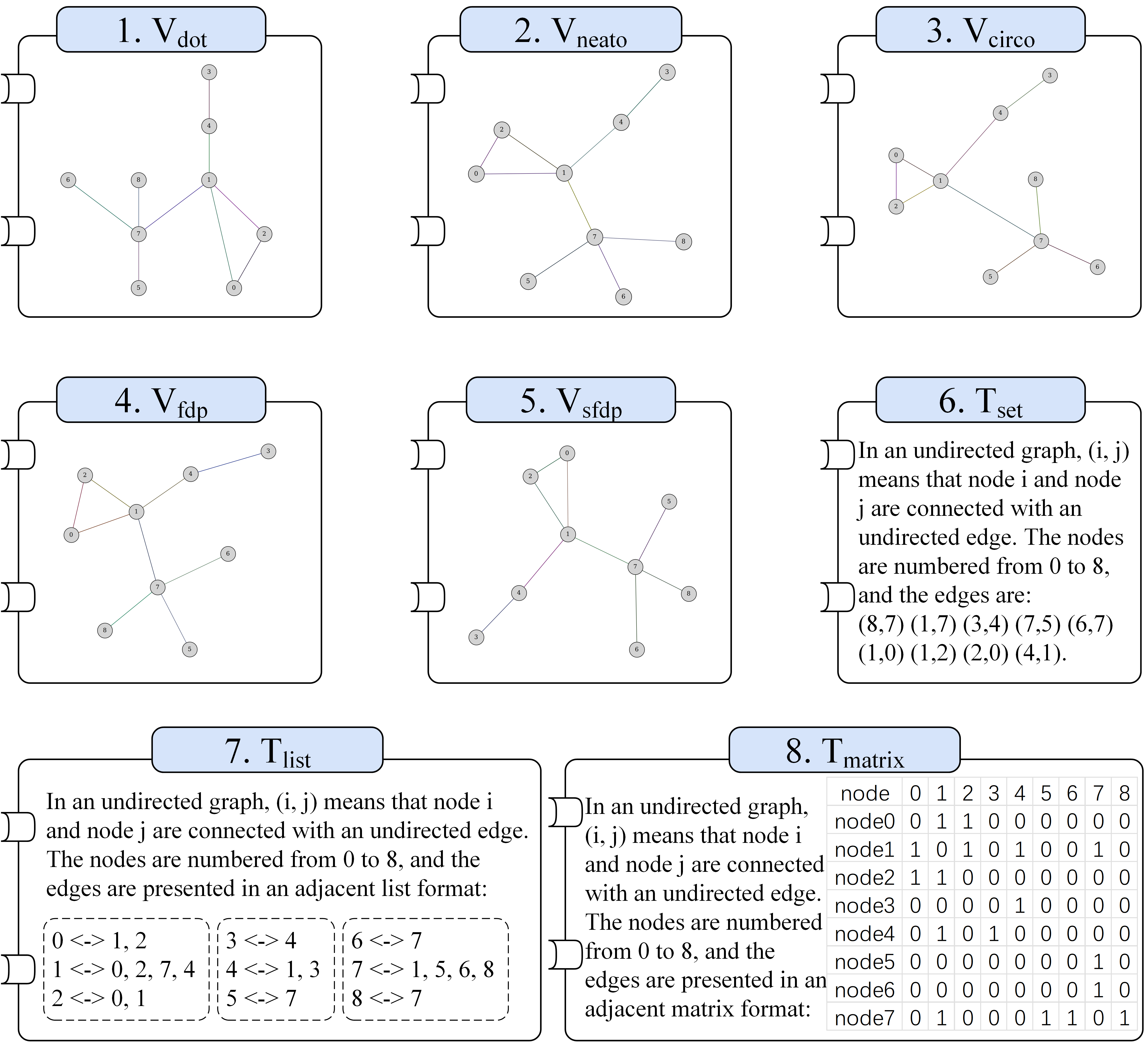}
\caption{An Illustration of eight candidate TRFs in $\mathcal{F}_{ZS}$.}
    \label{fig: examples}
\end{figure}
\begin{itemize}
    \item \textbf{$V_{dot}$}: This visual TRF adopts a hierarchical layout algorithm, which minimizes edge crossings and maintains edge directions (either top-to-bottom or left-to-right). It is particularly effective for representing organizational structures or flowcharts as a visual topology.
    
    \item \textbf{$V_{neato}$}: As a visual TRF, it employs a spring model \citep{fruchterman1991graph} layout algorithm. By simulating edges as springs and nodes as mutually repelling entities, it generates an optimal arrangement through force simulation, making it well-suited for visualizing undirected graphs and network relationships.
    
    \item \textbf{$V_{circo}$}: This visual TRF uses a circular layout algorithm that places nodes in a ring structure to reduce edge crossings. It excels at highlighting cyclic topologies, such as those in electronic circuits or advanced abstract syntax trees.
    
    \item \textbf{$V_{fdp}$}: The visual TRF also relies on the spring model but with an optimized algorithm for large graphs. While prioritizing speed, it may sacrifice some node distribution uniformity compared to neato, balancing efficiency and visual clarity for larger topologies.
    
    \item \textbf{$V_{sfdp}$}: This visual TRF incorporates a multi-level force-directed algorithm to extend the fdp layout algorithm to even larger-scale graphs. It maintains force-directed layout characteristics while handling thousands of nodes efficiently, making it ideal for visualizing large topological structures.
\end{itemize}
To align with the scope of our paper, we do not delve further into additional algorithmic details of these layout methods. For comprehensive implementation specifics of each layout algorithm, readers are referred to the Graphviz technical paper \citep{graphviz} and its corresponding Python package documentation. Notably, we have integrated a dedicated interface within our codebase to facilitate the generation of these visual TRFs, and we recommend that users utilize this interface for direct invocation.

\paragraph{More Details for Textual TRFs Generation}
The textual $T_{set}$, $T_{list}$, and $T_{mat}$ present the graph topology structure via Edge set, Adjacent List, and Adjacent Matrix, respectively. Their prompt templates are detailed in Table \ref{tab: prompt1} to \ref{tab: prompt3}. For $T_{set}$, the prompts are derived from \citep{NLGraph}, while the prompts for $T_{list}$ and $T_{mat}$ are designed and polished by ourselves. To be specific, we introduce these textual TRFs as follows:

\begin{itemize}
    \item \textbf{$T_{set}$}: This textual TRF represents graph topology through an unordered collection of edge tuples, where each tuple explicitly denotes a connection between two nodes (e.g., $(u, v)$ for an edge from node $u$ to node $v$). By omitting redundant structural framing, it achieves high information density, storing all edge relationships in a compact format. This characteristic makes it particularly efficient for tasks requiring quick enumeration of all connections, such as edge existence checks or basic subgraph extraction. However, its unordered nature may complicate reasoning about node neighborhoods or hierarchical relationships, as no implicit grouping of edges by source node is provided.
    
    \item \textbf{$T_{list}$}: Structured as a node-centric inventory, this textual TRF organizes edges by their source nodes, listing all adjacent target nodes for each node in a sequentially ordered format (sorted by node labels). This design strengthens the visibility of neighbor relationships, as all connections originating from a specific node are grouped together, facilitating tasks like neighbor counting, path traversal, or community detection. The sorted arrangement of nodes and edges (by label) further enhances readability, enabling systematic scanning of topological patterns. Compared to $T_{set}$, it introduces moderate structural overhead but significantly improves accessibility to node-specific relationship information.
    
    \item \textbf{$T_{mat}$}: This textual TRF encodes graph topology as a tabular matrix where rows and columns correspond to nodes, and cell values indicate edge presence (e.g., 1 for an existing edge, 0 for absence). While the inclusion of 0 values for non-edges introduces redundancy, especially in sparse graphs, where most cells are 0, it offers exceptional intuitiveness for visualizing global connectivity patterns. The matrix structure allows for immediate identification of symmetric relationships (via diagonal symmetry) and dense subgraphs (via contiguous blocks of 1s), making it well-suited for tasks like bipartite graph analysis or adjacency pattern recognition. Despite its lower information density, the grid-based format aligns with human cognitive patterns for spatial relationship processing, reducing the cognitive load for certain types of topological reasoning.
\end{itemize}

\begin{table*}[htbp]
  \footnotesize 
  \centering 
	\renewcommand{\arraystretch}{1.1}
\resizebox{\linewidth}{!}{
  \begin{tabular}
  {p{0.8in}|p{6.2in}} \toprule
 \textbf{Tasks}
		&  \textbf{Template}  \\
		\midrule
		Connectivity/\newline Cycle/ \newline Hamilton Path
		& \texttt{In an undirected graph, (i,j) means that node i and node j are connected with an undirected edge. The nodes are numbered from [P] to [P], and the edges are: \newline         ([P], [P]) , ([P], [P])...} 
	\\ \cline{1-2}
Topological Sort & \texttt{In a directed graph with [P] nodes numbered from [P] to [P]:\newline node [P] should be visited before node [P]\newline node [P] should be visited before node [P]\newline ...} 
 \\ \cline{1-2} 
Shortest Path & \texttt{In an undirected graph, the nodes are numbered from [P] to [P], and the edges are:\newline an edge between node [P] and node [P] with weight [P],\newline an edge between node [P] and node [P] with weight [P], \newline ...} 
 \\ \cline{1-2} 
Maximum Flow & \texttt{In a directed graph, the nodes are numbered from [P] to [P], and the edges are: \newline
an edge from node [P] to node [P] with capacity [P], \newline
an edge from node [P] to node [P] with capacity [P], \newline
...}
\\ \cline{1-2}
Bipartite Graph Matching & \texttt{There are [P] hosts numbered from [P] to [P], and [P] tasks numbered from [P] to [P]. Each host has a set of tasks that it is interested in: \newline Host [P] is interested in task [P].\newline Host [P] is interested in task [P].\newline ...} 
 \\  \hline
 Link Predict/ \newline Node Classify
		& \texttt{In an undirected graph, (i,j) means that node i and node j are connected with an undirected edge. The nodes are numbered from [P] to [P], and the edges are: \newline         ([P], [P]) , ([P], [P])...\newline The node attributes are: \newline Node [P], Attribute [P]\newline Node [P], Attribute [P] \newline ...}\\
\hline
  \end{tabular}}
    \caption{Prompt templates of $T_{set}$, where [P]s are placeholders that will be substituted for specific graph topology.}
    \label{tab: prompt1}
  \end{table*}

\begin{table*}[htbp]
  \footnotesize 
  \centering 
	\renewcommand{\arraystretch}{1.1}
\resizebox{\linewidth}{!}{
  \begin{tabular}
  {p{0.8in}|p{6.2in}} \toprule
 \textbf{Tasks}
		&  \textbf{Template}  \\
		\midrule
		Connectivity/\newline Cycle/ \newline Hamilton Path
		& \texttt{In an undirected graph, (i,j) means that node i and node j are connected with an undirected edge. The nodes are numbered from [P] to [P], and the edges are presented in an adjacent list format: \newline         [P] <-> [P], [P], [P],...
        \newline
        [P] <-> [P], [P], [P],...\newline...}
	\\ \cline{1-2}
Topological Sort & \texttt{In a directed graph with [P] nodes numbered from [P] to [P]:\newline node [P] should be visited before node [P], [P], [P],...\newline node [P] should be visited before node [P], [P], [P],...\newline ...} 
 \\ \cline{1-2} 
Shortest Path & \texttt{In an undirected graph, the nodes are numbered from [P] to [P], and the edges are presented in an adjacent list format:\newline node [P] is connected to: node [P] with distance: [P], node [P] with distance: [P],... \newline node [P] is connected to: node [P] with distance: [P], node [P] with distance: [P],... \newline ...} 
 \\ \cline{1-2} 
Maximum Flow & \texttt{In a directed graph, the nodes are numbered from [P] to [P], and the edges are presented in an adjacent list format:\newline node [P] is connected to: node [P] with capacity: [P], node [P] with capacity: [P],... \newline node [P] is connected to: node [P] with capacity: [P], node [P] with capacity: [P],... \newline ...}
\\ \cline{1-2}
Bipartite Graph Matching & \texttt{There are [P] hosts numbered from [P] to [P], and [P] tasks numbered from [P] to [P]. Each host has a set of tasks that it is interested in: \newline Host [P] is interested in tasks [P], [P], [P],....\newline Host [P] is interested in task [P], [P], [P],....\newline ...} 
 \\  \hline
 Link Predict/ \newline Node Classify
		& \texttt{In an undirected graph, (i,j) means that node i and node j are connected with an undirected edge. The nodes are numbered from [P] to [P], and the edges are presented in an adjacent list format: \newline         [P] <-> [P], [P], [P],...
        \newline
        [P] <-> [P], [P], [P],...\newline... \newline The node attributes are: \newline Node [P], Attribute [P]\newline Node [P], Attribute [P] \newline ...}\\
\hline
  \end{tabular}}
    \caption{Prompt templates of $T_{list}$, where [P]s are placeholders that will be substituted for specific graph topology.}
    \label{tab: prompt2}
  \end{table*}
\begin{table*}[htbp]
  \footnotesize 
  \centering 
	\renewcommand{\arraystretch}{1.1}
\resizebox{\linewidth}{!}{
  \begin{tabular}
  {p{0.8in}|p{6.2in}} \toprule
 \textbf{Tasks}
		&  \textbf{Template}  \\
		\midrule
		Connectivity/\newline Cycle/ \newline Hamilton Path
		& \texttt{In an undirected graph, (i,j) means that node i and node j are connected with an undirected edge. The nodes are numbered from [P] to [P], and the edges are represented in an adjacent matrix format \newline            :\ \  \ node0\ \ \ \ 1\ \ \ \ 2...
        \newline node0\ \ [P]\ \ [P]\ \ [P],...\newline node1\ \ [P]\ \ [P]\ \ [P],...\newline
        node2\ \ [P]\ \ [P]\ \ [P],...\newline
        ...}
	\\ \cline{1-2}
Topological Sort & \texttt{In a directed graph with [P] nodes numbered from [P] to [P], the edges are represented in an adjacent matrix format \newline            :\ \  \ node0\ \ \ \ 1\ \ \ \ 2...
        \newline node0\ \ [P]\ \ [P]\ \ [P],...\newline node1\ \ [P]\ \ [P]\ \ [P],...\newline
        node2\ \ [P]\ \ [P]\ \ [P],...\newline
        ...} 
 \\ \cline{1-2} 
Shortest Path/\newline Maximum Flow & \texttt{In an undirected graph, the nodes are numbered from [P] to [P], and the edges are represented in an adjacent matrix format with weights \newline            :\ \  \ node0\ \ \ \ 1\ \ \ \ 2...
        \newline node0\ \ [P]\ \ [P]\ \ [P],...\newline node1\ \ [P]\ \ [P]\ \ [P],...\newline
        node2\ \ [P]\ \ [P]\ \ [P],...\newline
        ...} 
 \\ \cline{1-2} 
Bipartite Graph Matching & \texttt{There are [P] hosts numbered from [P] to [P], and [P] tasks numbered from [P] to [P]. Each host has a set of tasks that it is interested in \newline            :\ \  \ Task0\ \ \ \ 1\ \ \ \ 2...
        \newline Host0\ \ [P]\ \ [P]\ \ [P],...\newline Host1\ \ [P]\ \ [P]\ \ [P],...\newline
        Host2\ \ [P]\ \ [P]\ \ [P],...\newline
        ...} 
 \\  \hline
 Link Predict/ \newline Node Classify
		& \texttt{In an undirected graph, (i,j) means that node i and node j are connected with an undirected edge. The nodes are numbered from [P] to [P], and the edges are represented in an adjacent matrix format \newline            :\ \  \ node0\ \ \ \ 1\ \ \ \ 2...
        \newline node0\ \ [P]\ \ [P]\ \ [P],...\newline node1\ \ [P]\ \ [P]\ \ [P],...\newline
        node2\ \ [P]\ \ [P]\ \ [P],...\newline
        ...\newline The node attributes are: \newline Node [P], Attribute [P]\newline Node [P], Attribute [P] \newline ...}\\
\hline
  \end{tabular}}
    \caption{Prompt templates of $T_{mat}$, where [P]s are placeholders that will be substituted for specific graph topology.}
    \label{tab: prompt3}
  \end{table*}

\section{B. TRFP Dataset Construction}
The TRFP dataset is designed to investigate query-specific preferences among TRFs within $\mathcal{F}_{ZS}$. To begin with, we generate 1K samples of QA for each task in \{Conn, Cyc, TP, SP, MF, BGM, HP\}, where the graph topologies are randomly generated by the Erdős–Rényi \citep{er_model} model ($\text{node count }N \in [3,30], \text{edge probability} \in [0.1,0.7]$). For SP and MF, the edge weights are randomly assigned $\in \{1,2,...,10\}$. Graphs that fail to meet task-specific validity criteria are regenerated (We make sure there exists at least a valid answer), ensuring we obtain 7K valid samples (1K per task).

Then, for each question, we evaluate every TRF in $\mathcal{F}_{ZS}$ by running $k$ independent trials to compute the corresponding GRE metric. TRFs are ranked for each question based on their GRE scores in descending order; in cases of tied GRE scores, the tied TRFs are treated as multi-label preferences. By pairing each question with its most preferred TRFs (those with the highest GRE scores), we construct the TRF Preference dataset, denoted as $\mathcal{D}_{\text{TRFP}} = \{(q_i, \mathcal{F}_{q_i}^*)\}$.

To validate response correctness, we use dedicated algorithms for each task:
\begin{itemize}
    \item Connectivity (Conn) identification answers are judged using the Union-Find algorithm \citep{tarjan1975efficiency}.
    \item Cycle detection (Cyc) results are verified via Depth-First Search (DFS) \citep{hopcroft1973algorithm}.
    \item Topological sorting (TS) answers are checked by verifying edge direction constraints in the sequence \citep{kahn1962topological}.
    \item Shortest path (SP) results are validated using Dijkstra's algorithm \citep{dijkstra2022note}.
    \item Maximum flow (MF) answers are judged via the Edmonds-Karp algorithm \citep{edmonds1972theoretical}.
    \item Bipartite graph matching (BGM) results are verified using the Hopcroft-Karp algorithm \citep{hopcroft1973n}.
    \item Hamiltonian path (HP) answers are checked using a backtracking-based approach \citep{bellman1962dynamic}.
\end{itemize}

The construction of $\mathcal{D}_{\text{TRFP}}$ is contextualized by both the target LMM and the specific definition of the GRE metric. Consequently, distinct LMMs require unique instantiations of the TRFP dataset, while adjustments to the $\alpha$ parameter, which modulates the trade-off between accuracy and brevity in necessitate corresponding updates to $\mathcal{D}_{\text{TRFP}}$ to reflect the shifted weightings in the metric.

\section{C. Label Distributions in TRFP datasets}
In this section, we present the complete label distributions of TRFP datasets for both GPT-4o Pro (Table \ref{tab: preference_4o}). According to this table, we have additional findings beyond those mentioned in the main body of the paper:
\begin{itemize}
    \item 1) When comparing GPT-4o and Gemini-2.5 Pro, they exhibit highly consistent task-specific TRF rankings, indicating the task-specific TRF preferences can go beyond model biases.
    \item 2) Specifically, for Conn and Cyc, the suboptimal textual TRFs $(T_{list}, T_{mat}, T_{set})$ also have a consistent inner priority across both models, with $T_{list} > T_{mat} > T_{set}$, highlighting the relative priority of textual TRFs in expressing topological connectivity effectively.
    \item 3) In the case of GPT-4o, even in tasks where other textual TRFs like $T_{list}$ and $T_{set}$ are preferred, $T_{mat}$ may be suboptimal, indicating a model-specific misalignment with this TRF form. However, this discrepancy is not observed in Gemini-2.5 Pro, suggesting that when someone intends to use $T_{mat}$ with its more intuitive graph adjacency representation, it is preferable to use Gemini-2.5 Pro over GPT-4o.
    \item 4) Analysis of the frequency gaps among the TRFs reveals that the preferences in Conn and Cyc are more pronounced, with larger variations in frequency. This is followed by BGM, and then TS, SP, MF, and HP, where the frequency gaps are more moderate.   
\end{itemize}

\begin{table*}[h!tbp]
    \centering
    \setlength{\tabcolsep}{3pt}
    \renewcommand{\arraystretch}{1.3} 
    \small 
    \begin{tabular}{llccccccc}
        \hline
        Model & TRFs & Conn & Cyc & TS & SP & MF & BGM & HP \\
        \hline
        \multirow{8}{*}{\makecell[c]{GPT-4o}} & 1st 
        &\cellcolor{gray!20}$V_{fdp}$ (92.3\%) 
        &\cellcolor{gray!20} $V_{sfdp}$ (85.1\%)
        & $T_{set}$ (58.5\%)
        & $T_{set}$ (19.5\%)
        & $T_{list}$ (21.7\%)
        &\cellcolor{gray!20} $V_{dot}$ (34.2\%)
        &$T_{list}$ (30.4\%)\\
        \cline{2-9}
        & 2nd & 
        \cellcolor{gray!20}$V_{neato}$ (92.1\%) 
        & \cellcolor{gray!20}$V_{fdp}$ (12.9\%)
        & $T_{list}$ (36.2\%)
        &\cellcolor{gray!20} $V_{neato}$ (18.7\%)
        & $T_{set}$ (20.8\%)
        &\cellcolor{gray!20} $V_{circo}$ (20.3\%)
        & $T_{set}$ (20.3\%)\\ 
        \cline{2-9}
        & 3rd 
        &\cellcolor{gray!20} $V_{sfdp}$ (91.7\%)
        &\cellcolor{gray!20} $V_{dot}$ (12.2\%)
        &\cellcolor{gray!20} $V_{dot}$ (23.1\%)& $T_{list}$ (17.1\%)& $T_{mat}$ (16.7\%)
        &\cellcolor{gray!20} $V_{neato}$ (19.8\%)
        &\cellcolor{gray!20} $V_{circo}$ (18.8\%)\\

        \cline{2-9} & 4th 
        &\cellcolor{gray!20}$V_{circo}$ (84.4\%) 
        &\cellcolor{gray!20} $V_{circo}$ (10.5\%)
        & $T_{mat}$ (0.8\%)
        & \cellcolor{gray!20}$V_{fdp}$ (16.3\%)
        & \cellcolor{gray!20}$V_{neato}$ (10.8\%)
        &\cellcolor{gray!20} $V_{sfdp}$ (13.4\%)
        &\cellcolor{gray!20}$V_{dot}$ (17.4\%)\\
        \cline{2-9}
        & 5th & 
        \cellcolor{gray!20}$V_{dot}$ (73.0\%) 
        & \cellcolor{gray!20}$V_{fdp}$ (61.1\%)
        & \cellcolor{gray!20}$V_{sfdp}$ (0.8\%)
        &\cellcolor{gray!20} $V_{circo}$ (13.8\%)
        & \cellcolor{gray!20}$V_{fdp}$ (8.3\%)
        &\cellcolor{gray!20} $V_{fdp}$ (9.1\%)
        & \cellcolor{gray!20}$V_{sfdp}$ (15.9\%)\\ 
        \cline{2-9}
        & 6th 
        & $T_{list}$ (27.0\%)
        & $T_{list}$ (10.8\%)
        &\cellcolor{gray!20} $V_{neato}$ (0.8\%)& \cellcolor{gray!20}$V_{sfdp}$ (12.2\%)& \cellcolor{gray!20}$V_{dot}$ (7.5\%)
        &$T_{list}$ (7.0\%)
        &\cellcolor{gray!20} $V_{neato}$ (14.5\%)\\
        \cline{2-9} & 7th 
        &$T_{mat}$ (58.0\%) 
        &$T_{mat}$ (0.2\%)
        & \cellcolor{gray!20} $V_{circo}$ (0.8\%) 
        & \cellcolor{gray!20}$V_{dot}$ (10.6\%)
        & \cellcolor{gray!20}$V_{circo}$ (7.5\%)
        &  $T_{set}$ (1.6\%)
        & $T_{mat}$ (7.2\%)
        \\
        \cline{2-9}
        & 8th & $T_{set}$ (47.7\%) 
        & $T_{set}$ (0.0\%) 
        & \cellcolor{gray!20} $V_{fdp}$ (0.8\%)
        & $T_{mat}$ (6.5\%)
        & \cellcolor{gray!20}$V_{sfdp}$ (6.7\%)

        & $T_{mat}$ (0.5\%)
        & \cellcolor{gray!20}$V_{fdp}$ (0.0\%)\\ 
        \cline{2-9}
        \hline
        \multirow{8}{*}{\makecell[c]{Gemini \\ 2.5 Pro}} & 1st 
        &\cellcolor{gray!20} $V_{neato}$ (88.8\%)
        &\cellcolor{gray!20}
        $V_{neato}$ (97.2\%)
        &
        $T_{set}$ (41.0\%)
        &
        $T_{list}$ (48.6\%)
        &
        $T_{mat}$ (40.0\%)
        &\cellcolor{gray!20}
        $V_{fdp}$ (24.4\%)
        &
        $T_{list}$ (42.9\%)\\
        \cline{2-9}
        & 2nd &\cellcolor{gray!20}
        $V_{fdp}$ (88.2\%)
        & \cellcolor{gray!20} $V_{sfdp}$ (93.0\%)
        & $T_{list}$ (30.3\%)
        & $T_{mat}$ (37.6\%)
        & $T_{list}$ (36.0\%)
        & \cellcolor{gray!20} $V_{sfdp}$ (14.5\%)
        & $T_{set}$ (31.7\%)
        \\
        \cline{2-9}
        & 3rd 
        &\cellcolor{gray!20} $V_{sfdp}$ (88.2\%)
        & \cellcolor{gray!20} $V_{dot}$ (71.8\%)
        & $T_{mat}$ (15.4\%)
        & $T_{set}$ (26.6\%)
        & $T_{set}$ (14.0\%)
        & \cellcolor{gray!20} $V_{neato}$ (14.5\%)
        & $T_{mat}$ (30.2\%)\\
        \cline{2-9} 
        \cline{2-9} & 4th 
        &\cellcolor{gray!20}$V_{circo}$ (87.0\%) 
        &\cellcolor{gray!20} $V_{circo}$ (71.1\%)
        & \cellcolor{gray!20}$V_{dot}$ (6.9\%)
        & \cellcolor{gray!20}$V_{sfdp}$ (13.8\%)
        & \cellcolor{gray!20}$V_{sfdp}$ (6.0\%)
        &\cellcolor{gray!20} $V_{dot}$ (13.7\%)
        &\cellcolor{gray!20}$V_{fdp}$ (9.5\%)\\
        \cline{2-9}
        & 5th & 
        \cellcolor{gray!20}$V_{dot}$ (83.4\%) 
        & \cellcolor{gray!20}$V_{neato}$ (8.3\%)
        & \cellcolor{gray!20}$V_{fdp}$ (4.3\%)
        &\cellcolor{gray!20} $V_{dot}$ (11.0\%)
        & \cellcolor{gray!20}$V_{neato}$ (2.0\%)
        &$T_{mat}$ (12.2\%)
        & \cellcolor{gray!20}$V_{circo}$ (9.5\%)\\ 
        \cline{2-9}
        & 6th 
        & $T_{list}$ (61.3\%)
        & $T_{list}$ (0.5\%)
        &\cellcolor{gray!20} $V_{neato}$ (2.1\%)& \cellcolor{gray!20}$V_{fdp}$ (8.3\%)& \cellcolor{gray!20}$V_{circo}$ (2.0\%)
        & \cellcolor{gray!20}$V_{circo}$ (9.2\%)
        &\cellcolor{gray!20} $V_{sfdp}$ (9.6\%)\\
        \cline{2-9} & 7th 
        &$T_{mat}$ (8.5\%) 
        &$T_{mat}$ (5.2\%)
        & \cellcolor{gray!20} $V_{sfdp}$ (0.0\%) 
        & \cellcolor{gray!20}$V_{circo}$ (4.6\%)
        & \cellcolor{gray!20}$V_{dot}$ (0.0\%)
        &  $T_{set}$ (6.9\%)
        & $V_{dot}$ (0.0\%)
        \\
        \cline{2-9}
        & 8th & $T_{set}$ (5.1\%) 
        & $T_{set}$ (2.1\%) 
        & \cellcolor{gray!20} $V_{fdp}$ (0.8\%)
        & \cellcolor{gray!20}$V_{neato}$ (2.8\%)
        & \cellcolor{gray!20}$V_{fdp}$ (0.0\%)

        & $T_{list}$ (5.3\%)
        & \cellcolor{gray!20}$V_{fneato}$ (0.0\%)\\ 
        \cline{2-9}
        \hline
    \end{tabular}
    \caption{Complete ranking of TRFs with respect to their label frequency in the TRFP dataset of GPT-4o and Gemini-2.5 Pro. By differing visual or textual TRFs with colors, \textbf{the special preference patterns of tasks are explicitly exposed}.}
    \label{tab: preference_4o}
\end{table*}

\section{D. Proof}
We here provide the theorem with proof to demonstrate the pareto optimality of ideal dynamic routing. 

\begin{theorem}[GRE-based Dynamic Routing Pareto Optimality]
For any question distribution $\mathcal{D}_q$ and tradeoff parameter $\alpha>0$, the optimal router $R^*$ satisfying
$\forall f\in\mathcal{F}_{ZS}, R^*\succ R_f$, where $R^*$ always select a TRF $f_q^{R^*} \in \mathcal{F}^*_q$, and  $R_f$ represents the routing always select $f$.
The strict inequality establishes when $f$ is suboptimal for any $q\in\text{supp}(\mathcal{D}_q)$.
\end{theorem}

\begin{proof}
Given $E^k_f(q) = \text{Acc}^k_f(q) + \alpha\text{Eff}^k_f(q)$. For any fixed mode $f$:
\begin{align*}
\mathbb{E}_q[E^k_{f^{R*}_q}(q)] &= \mathbb{E}_q\left[\max_{m'\in\mathcal{M}} {E}^k_{m'}(q)\right] \\
&\geq \mathbb{E}_q[{E}^k_f(q)] \quad \text{(pointwise optimality)} \\
&= \mathbb{E}_q[\text{Acc}^k_f(q)] + \alpha\mathbb{E}_q[\text{Eff}^k_f(q)]
\end{align*}
Rearranging terms, we have $\mathbb{E}_q[\text{Acc}_{f^{R*}_q}^k(q)] - \mathbb{E}_q[\text{Acc}^k_f(q)]  
 \geq \alpha\left(\mathbb{E}_q[\text{Eff}^k_f(q)] - \mathbb{E}_q[\text{Eff}_{f^{R*}_q}^k(q)]\right)$. If $\mathbb{E}_q[\text{Eff}^k_{f^{R*}_q}(q)] < \mathbb{E}_q[\text{Eff}^k_f(q)]$, the RHS becomes positive, forcing $\mathbb{E}_q[\text{Acc}^k_{f^{R*}_q}(q)] > \mathbb{E}[\text{Acc}^k_f(q)]$. Otherwise $\mathbb{E}_q[\text{Eff}^k_{f^{R*}_q}(q)] \geq \mathbb{E}_q[\text{Eff}^k_f(q)]$ directly holds. Thus $R^*$ either strictly improves accuracy while matching efficiency, or maintains accuracy while strictly improving efficiency. This establishes Pareto dominance over any fixed mode $m$.
\end{proof}
\section{E. Data and Task Details}
\subsection{Task Introduction}
We introduce Tasks in this section. The 7 in-domain graph algorithmic QA tasks include:
\begin{itemize}
\item \textbf{Connectivity}~\citep{sedgewick2001algorithms} (abbreviated as Conn): Assess whether two randomly chosen nodes $u$ and $v$ in an undirected graph are linked.

\item \textbf{Cycle}~\citep{sedgewick2001algorithms} (abbreviated as Cyc): Determine if there is a cycle present within an undirected graph.

\item \textbf{Topological Sort}~\citep{kahn1962topological} (abbreviated as TS): Identify a valid topological ordering for a directed acyclic graph. This sort provides a sequence of nodes such that for every directed edge $u \leftarrow v$, node $u$ precedes node $v$ in the sequence.

\item \textbf{Shortest Path}~\citep{dijkstra2022note} (abbreviated as SP): Locate the shortest route between two nodes in a weighted undirected graph. The shortest path is defined as the route with the smallest total edge weight connecting the two nodes.

\item \textbf{Maximum Flow}~\citep{ford1956maximal} (abbreviated as MF): Compute the maximum flow from a source node to a destination node in a network graph.

\item \textbf{Bipartite Graph Matching}~\citep{karp1990optimal} (abbreviated as BGM): Identify the largest matching set in a bipartite graph. A matching set consists of edges where no two edges share a common node.

\item \textbf{Hamilton Path}~\citep{gould2003advances} (abbreviated as HP): Discover a Hamiltonian path in an undirected graph. This path visits each node exactly once.

\end{itemize}
The 2 out-of-domain downstream application tasks are:
\begin{itemize}
    \item \textbf{Link Prediction}: Predict whether a link or edge exists between two nodes in a network, based on the current structure and attributes of the graph. It is one of the cornerstone tasks in graph learning and has wide applications in social network analysis \citep{social1}, recommendation systems \citep{lightgcn}, and biological network studies \citep{drug1}.

    \item \textbf{Node Classification}: This task is concerned with predicting the categories of nodes within a graph.  Node classification is widely used in applications such as community detection \citep{newman2004finding}, fraud detection \citep{akoglu2015graph}, and identifying roles in social networks \citep{wasserman1994social}.
\end{itemize}
\subsection{Dataset Statistics}
We adopt the GVLQA-BASE \citep{wei2024gita} benchmark for in-domain tasks.  For out-of-domain applications, we adopt the ca-GrQC and ca-HepTh \citep{cadataset} datasets for link prediction (LP), and use the PolBlog \citep{adamic2005political} and Cora \citep{yang2016revisiting} datasets for the node classification (NC) task. Data statistics are provided in Table \ref{tab: data1} and \ref{tab: data2}.

\begin{table*}[!htbp]
\centering
\begin{tabular}{lccccccc}
\toprule
 & Conn & Cyc & TS & SP & MF & BGM & HP \\ 
\midrule
\#sample & 16,410 &  4,100 &  2,910 &  1,560 &  1,500 & 1,860 & 900 \\
\#nodes & 25.01 & 23.42 & 21.86 & 13.65 & 13.90 & 21.13 & 13.24 \\ 
\#edges & 95.46 & 23.66 & 114.10 & 23.99 & 49.16 & 51.03 & 45.05 \\ \bottomrule
\end{tabular}
\caption{Data Statistics for in-domain graph algorithmic QA tasks.}
\label{tab: data1}
\end{table*}

\begin{table*}[!hbtp]
\centering
\resizebox{0.7\textwidth}{!}{%
    \begin{tabular}{cccccc}
    \toprule
     & ca-GrQC & ca-HepTh & PolBlogs & Cora & CiteSeer\\ 
     \midrule
    \# Nodes & 5,242 & 9,877 & 1,490 & 2,708 & 3,327\\
    \# Edges & 14,496 & 25,998 & 19,025 & 5,278 & 4,676\\
    domain  & collaboration  & collaboration & social & citation  & citation\\
    average degree & 5.53 & 5.26 & 25.54 & 3.9  & 2.74\\
    
    \bottomrule
    \end{tabular}%
}
\caption{Data Statistics for downstream applications}
\label{tab: data2}
\end{table*}

\section{F. Implementation Details}
In our experiment, we use DeBERTaV3-base as TRF Router, which is trained with learning rate $\in \{5e-5, 5e-6\}$, weight decay $\in \{1e-2, 1e-3\}$, epoch $\in \{6, 8, 10\}$, batch size $\in \{16, 32, 64\}$. All experiments are conducted on a single NVIDIA A100 GPU. For all methods in comparison, the task instructions are unified in Table \ref{tab: instruction}, which is followed by a control instruction (provided in \ref{tab: control}) to specify the place and format of the answer to be extracted. More implementation details for each method are as follows:
\begin{itemize}
    \item GPT-4o/Gemini-2.5 Pro: For the backbone LMMs, we examine their graph QA performances by invoking the official API Interface to directly answer the questions with instructions. Due to the prevalent nature of edge set in existing works, we adopt $T_{set}$ to represent the graph topology during implementation.
    \item Vanilla Chain-of-thought (CoT) \citep{wei2022chain}: CoT has demonstrated that the simple prompts that drive the LMMs in analytical, step-by-step thinking can boost the general range of capabilities of LMM reasoning. We adopt CoT in our baseline by appending a sentence of CoT prompt in the input tail: ``\texttt{Please think step by step and present the rationales in a well-structured manner, to make the answer more reliable and robust.}''. Due to the prevalent nature of edge set in existing works, we adopt $T_{set}$ to represent the graph topology during implementation.
    \item NLGraph \citep{NLGraph}/ GraphDPR \citep{li2024visiongraph}/ GITA \citep{wei2024gita}: NLGraph proposes to use BAG prompting to conceptualize a graph and algorithmic prompting to specify the algorithm. GraphDPR proposes to employ the external graph algorithmic toolkits to generate intermediate descriptions and code, which are utilized to enhance multi-step reasoning. GITA simultaneously adopts a visual TRF and (similar to $V_{dot}$) a textual TRF (Similar to $T_{set}$) to solve graph algorithmic problems by leveraging the complementary nature of the visual and textual mutual-enhancement across tasks. We evaluate these methods based on their own official implementations, but modifying the prompt templates to that we have specified in Table \ref{tab: prompt1}, \ref{tab: instruction}, and \ref{tab: control}.
\end{itemize}

\begin{table*}[htbp]
  \footnotesize 
  \centering 
	\renewcommand{\arraystretch}{1.1}
\resizebox{\linewidth}{!}{
  \begin{tabular}
  {p{0.8in}|p{6.2in}} \toprule
 \textbf{Tasks}
		&  \textbf{Task Instruction}  \\
		\midrule
		Connectivity
		& \texttt{Is there a path between node [P] and node [P] in this undirected graph?} 
	\\ \cline{1-2}
    Cycle
		& \texttt{Is there a cycle in this undirected graph?} 
	\\ \cline{1-2}

Topological Sort & \texttt{This representation depicts a directed graph, in which each directed edge from node A to node B signifies that, according to the topological order, node A must precede node B. Q: The topological order of the directed graph is:} 
	\\ \cline{1-2}
Shortest Path & \texttt{This representation illustrates a directed graph, with each edge's capacity indicated by a numerical label in close proximity.Q: What is the maximum flow from node 4 to node 0:} 
 \\ \cline{1-2} 
Maximum Flow & \texttt{This representation illustrates a directed graph, with each edge's capacity indicated by a numerical label in close proximity. Q: What is the maximum flow from node [P] to node [P]:}
\\ \cline{1-2}
Bipartite Graph Matching & \texttt{There are [P] hosts numbered from [P] to [P], and [P] tasks numbered from [P] to [P]. Each host has a set of tasks that it is interested in, represented by arrows from a host to a task in the diagram. However, each host is capable of solving only one task, and similarly, each task can be resolved by just one host. Q:  What is the maximum number of hosts that can be assigned a task they are interested in?} 
 \\  \hline
Hamilton Path & \texttt{Q: Begin with node 0, what is the path in this graph that visits every node exactly once?} \\
\hline
 Link Predict
		& \texttt{The task is link prediction, aiming to predict the presence or absence of an unknown edge between Node [P] and Node [P] based on the known graph structure. Q: Does an unknown edge exist between Node [P] and Node [P]?}\\

\hline
 Node Classify
		& \texttt{The task is semi-supervised node classification, and needs to predict which class Node [P] belongs to, based on graph structure and known node classes. Q: Node [P] belongs to Class: Note that capacity is directional, allowing flow only in the edge direction; reverse edge direction should not be considered in the path. }\\
\hline
  \end{tabular}}
    \caption{Unified Task Instructions in experimental comparisons, where [P]s are placeholders that will be substituted in specific questions.
    }
    \label{tab: instruction}
  \end{table*}

\begin{table*}[htbp]
  \footnotesize 
  \centering 
	\renewcommand{\arraystretch}{1.1}
\resizebox{\linewidth}{!}{
  \begin{tabular}
  {p{0.8in}|p{6.2in}} \toprule
 \textbf{Tasks}
		&  \textbf{Task Instruction}  \\
		\midrule
		Connectivity/\newline Cycle/\newline Link Predict
		& \texttt{Please put the answer between <answer> and </answer> tags. For example, <answer>Yes</answer> or <answer>No</answer>.} 
	\\ \cline{1-2}

Topological Sort/\newline Shortest Path/\newline Hamilton Path& \texttt{Please put the answer between <answer> and </answer> tags. For example, <answer>0->1->2->3->4</answer> or <answer>0->1->3->7->8->4->6->5->9->2</answer>. } 
	\\ \cline{1-2}
Maximum Flow & \texttt{Please put the answer between <answer> and </answer> tags. For example, <answer>3</answer> or <answer>8</answer>. } 
 \\  \hline
Node Classify & \texttt{Please put the answer between <answer> and </answer> tags. For example, <answer>Class 1</answer> or <answer>Class 3</answer>.} \\
\hline
  \end{tabular}}
    \caption{Control Task Instructions in experimental comparisons.
}
    \label{tab: control}
  \end{table*}

\section{G. TRF Router versus single TRFs (both GPT-4o and Gemini-2.5 Pro)}

To demonstrate the importance of the proposed TRF Routing, we present metrics comparing individual TRFs with the TRF Router within the DynamicTRF framework. Additionally, we include the performance of \textit{`Ideal Routing'}, where the TRF with the optimal Global Routing Efficiency (GRE) is always selected, representing the upper bound capability of TRF routing.

While we have already presented such analysis for GPT-4o in our manuscript in section 5.3 based on Table 5, it is important to note that Table 5 only contains results for GPT-4o. Here, we provide a complete comparison table including the results for Gemini-2.5 Pro in Table \ref{tab: abl2}. The observations for Gemini-2.5 Pro are similar to those we have reported in section 5.3 for GPT-4o, indicating that there is no single dominant TRF across all tasks. The top TRFs for each task align closely with the task preferences shown in the TRFP dataset (Table \ref{tab: preference_4o}).

With the TRF Router, DynamicTRF achieves the highest GRE scores across all tasks compared to individual TRFs, showcasing its superior performance in balancing accuracy and efficiency. However, there still exists a gap between the TRF Router and `Ideal Routing', emphasizing the continued potential of TRF routing.

\begin{table*}[htbp]
    \centering
    \setlength{\tabcolsep}{0.3pt}
    \renewcommand{\arraystretch}{1.1} 
    \small 
    \begin{tabular}{lccccccccccccccc}
        \hline
        & \multicolumn{2}{c}{\raisebox{-0.6ex}[0pt]{Conn}} 
        & \multicolumn{2}{c}{\raisebox{-0.6ex}[0pt]{Cyc}} 
        & \multicolumn{2}{c}{\raisebox{-0.6ex}[0pt]{TS}} 
        & \multicolumn{2}{c}{\raisebox{-0.6ex}[0pt]{SP}} 
        & \multicolumn{2}{c}{\raisebox{-0.6ex}[0pt]{MF}} 
        & \multicolumn{2}{c}{\raisebox{-0.6ex}[0pt]{BGM}} 
        & \multicolumn{2}{c}{\raisebox{-0.6ex}[0pt]{HP}} \\
        TRF
        &\multicolumn{2}{c}{\raisebox{0.5ex}{\rule{2.05cm}{0.4pt}}}
        &\multicolumn{2}{c}{\raisebox{0.5ex}{\rule{2.05cm}{0.4pt}}}
        &\multicolumn{2}{c}{\raisebox{0.5ex}{\rule{2.05cm}{0.4pt}}}
        &\multicolumn{2}{c}{\raisebox{0.5ex}{\rule{2.05cm}{0.4pt}}}
        &\multicolumn{2}{c}{\raisebox{0.5ex}{\rule{2.05cm}{0.4pt}}}
        &\multicolumn{2}{c}{\raisebox{0.5ex}{\rule{2.05cm}{0.4pt}}}
        &\multicolumn{2}{c}{\raisebox{0.5ex}{\rule{2.05cm}{0.4pt}}}
        \\
                 & Acc(Tok) & GRE & Acc(Tok) & GRE & Acc(Tok) & GRE & Acc(Tok) & GRE & Acc(Tok) & GRE & Acc(Tok) & GRE & Acc(Tok) & GRE \\
        \midrule
        \multicolumn{15}{c}{\textit{GPT-4o}} \\
        \midrule
        $V_{dot}$
        &78.5(8.4) & 27.1
        &80.0(\underline{8.0}) & {28.3}
        &12.3(346.0) & 0.7
        &14.5(\underline{146.3}) & 1.2
        &7.5(410.2) & 0.4
        &\underline{91.2}(\underline{244.3}) & \underline{5.8}
        &13.3(\textbf{40.0}) & 2.1
        \\
        $V_{neato}$
        &95.1(\textbf{7.7}) & \textbf{34.3}
        &\underline{87.1}(\underline{8.0}) & \underline{30.8}
        &3.0(\textbf{30.0}) & 0.5
        &20.0(\textbf{130.1}) & 1.8
        &10.8(387.4) & 0.5
        &87.2(280.0) & 5.2
        &11.1(\textbf{40.0}) & 1.8
        \\
        $V_{circo}$
        &89.7(8.2) & 31.3
        &70.7(\underline{8.0}) & 25.0
        &3.0(\textbf{30.0}) & 0.5
        &18.2(153.5) & 1.5
        &8.1(421.3) & 0.4
        &88.2(271.6) & 5.3
        &14.4(\textbf{40.0}) & 2.3
        \\
        $V_{fdp}$
        &\underline{96.0}(\underline{8.1}) & \underline{33.6}
        &60.5(8.0) & 21.4
        &3.0(\textbf{30.0}) & 0.5
        &17.3(150.5) & 1.4
        &8.1(389.2) & 0.4
        &82.9(295.3) & 4.8
        &4.4(74.0) & 0.5
        \\
        $V_{sfdp}$
        &94.8(\underline{8.1}) & 33.3
        &85.1(\textbf{7.0}) & \textbf{32.2}
        &7.0(120.0) & 0.6
        &25.5(168.7) & 2.0
        &8.1(386.1) & 0.4
        &84.0(302.9) & 4.8
        &12.2(40.0) & 1.9
        \\
        $T_{set}$
        & 92.5(273.3) & 5.6
        & 52.7(480.6) & 2.4
        & \underline{36.6}(224.2) & \underline{2.4}
        & 54.6(566.0) & 2.3
        & \underline{25.3}(\textbf{362.9}) & \underline{1.3}
        & 69.5(370.1) & 3.6
        & \underline{50.0}(124.9) & 4.5
        \\
        $T_{list}$ 
        & 89.0(218.2) & 6.0
        & 53.2(359.7) & 2.8
        & 34.2(206.3) & 2.4
        & \underline{55.5}(518.9) & \underline{2.4}
        & 24.2(400.7) & 1.2
        & 65.8(410.8) & 3.2
        & 50.0(107.0) & \underline{4.8}
        \\
        $T_{mat}$
        &79.9(233.7) & 5.2
        &52.7(417.1) & 2.6
        &6.8(378.1) & 0.4
        &34.5(591.8) & 1.4
        &17.2(402.9) & 0.9
        &60.4(407.3) & 3.0
        &31.1(160.9) & 2.5
        \\
        \rowcolor{gray!20}  TRF Router
        & \textbf{96.1}(38.8) & 15.4 
        & \textbf{89.3}(75.9)& 10.3
        & \textbf{41.4}(176.1)& \textbf{3.1}
        & \textbf{68.4}(499.1)& \textbf{3.1}
        & \textbf{36.6}(\underline{385.2})& \textbf{1.9}
        & \textbf{92.0}(\textbf{233.6})& \textbf{6.0}
        & \textbf{61.1}(76.3)& \textbf{7.0}
        \\
        \midrule
        \textit{Ideal Routing}
        &\textit{100(7.9)} & \textit{35.6}
        &\textit{100(7.1)} & \textit{37.6}
        &\textit{44.5(268.0)} & \textit{2.7}
        &\textit{81.8(223.4)} & \textit{5.5}
        &\textit{53.8(380.0)} & \textit{2.8}
        &\textit{100(181.7)} & \textit{7.4}
        &\textit{76.7(72.2)} & \textit{9.0}
        \\
        \hline
        \multicolumn{15}{c}{\textit{Gemini-2.5 Pro}} \\
        \midrule
        $V_{dot}$
        &94.1(\underline{8.4}) & 32.5
        &72.5(\textbf{8.0}) & 25.6
        &23.2(1157.0) & 0.7
        &46.8(846.0) & 1.6
        &14.6(1113.5) & 0.4
        &93.9(1023.7) & 2.9
        & 30.6(1196.0) & 0.9
        \\
        $V_{neato}$
        & 99.6(\textbf{7.7}) & \textbf{35.8}
        &97.9(\textbf{8.0}) &\textbf{34.6}
        &7.9(\underline{885.0}) & 0.3
        & 53.2(866.9) & 1.8
        &25.0(1173.3) &0.7
        &96.5(1005.7) &3.0
        &27.4(1056.0) &0.8
        \\
        $V_{circo}$
        & 98.4(9.5) & 31.9
        &71.1(\textbf{8.0}) &25.1
        &3.0(1106.0) &0.1
        &40.4(852.8)&1.4
        &8.3(1056.4) &0.3
        &91.3(1137.7) &2.7
        &226.0(\textbf{30.0}) & \underline{4.1}
        \\
        $V_{fdp}$
        &\underline{99.6}(997.0) &32.0
        &61.0(\textbf{8.0}) &21.6
        &6.1(\textbf{638.0}) & 0.2
        &44.0(856.8) & 1.5
        &18.8(1158.0) & 0.6
        &\underline{98.3}(\underline{948.2}) &\underline{3.2}
        &21.0(\textbf{30.0}) &3.8\\
        $V_{sfdp}$
        &99.4(\underline{8.4}) &\underline{34.4}
        &93.7(8.0)&\underline{33.1}
        &5.5(1585.0) &0.1
        &46.8(828.7) &1.6
        &16.7(1178.7) &0.5
        &95.7(1042.2) & 3.0
        &30.6(891.0) & 1.0
        \\
        $T_{set}$
        &97.2(218.7) & 6.6
        &\underline{98.6}(716.6) & 3.7
        &\underline{84.1}(1395.9) & \underline{2.3}
        &93.6(810.8) & 3.3
        &91.7(1154.9) & 2.7
        &97.1(1076.8) & 3.0
        &96.8(678.1) & 3.7
        \\
        $T_{list}$ 
        &97.4(155.3) & 7.8
        & 96.5(652.8) & 3.8
        & 76.8(1532.3) & 2.0
        & \textbf{96.3}(\textbf{601.0}) & \textbf{3.9}
        & 92.2(\underline{1035.8}) & 2.9
        & 97.6(1129.9) & 2.9
        & 97.0(646.1) & 3.8
        \\
        $T_{mat}$
        &97.6(176.1) & 7.4
        &96.5(720.8) & 3.6
        &68.9(1554.7) & 1.7
        & 93.6(\underline{728.7}) & \underline{3.5}
        & \underline{95.8}(1042.6) & \underline{3.0}
        & 96.9(1033.8) & 3.0
        & \underline{98.4}(737.0) & 3.6\\
        \rowcolor{gray!20}  TRF Router
        &\textbf{100}(\textbf{12.9}) & 27.8
        &\textbf{99.3}(16.7) & 24.3
        &\textbf{87.8}(1191.2) & \textbf{2.5}
        &\textbf{96.3}(798.6) & 3.4
        &\textbf{100}(\textbf{1004.8}) & \textbf{3.2}
        &\textbf{100}(\textbf{776.0}) & \textbf{3.6}
        &\textbf{100}(254.6) & \textbf{6.3}
        \\
        \midrule
        \textit{Ideal Routing}
        &\textit{100(7.9)} & \textit{35.6}
        &\textit{100(7.1)} & \textit{37.6}
        &\textit{44.5(268.0)} & \textit{2.7}
        &\textit{81.8(223.4)} & \textit{5.5}
        &\textit{53.8(380.0)} & \textit{2.8}
        &\textit{100(181.7)} & \textit{7.4}
        &\textit{76.7(72.2)} & \textit{9.0}
        \\
        \hline
    \end{tabular}
    
    \caption{Comparison of the DynamicTRF framework with TRF Router versus single TRFs on in-domain tasks. `\textit{Ideal Routing}' means the routing is always to the best TRF.}
    \label{tab: abl2}
\end{table*}
\newpage
